\documentclass[10pt,twocolumn,letterpaper]{article}

\usepackage{cvpr}              

\usepackage{arydshln}
\usepackage{graphicx}
\usepackage{amsmath}
\usepackage{amssymb}
\usepackage{booktabs}
\usepackage{bm}
\usepackage{multirow}
\usepackage{mathrsfs}
\usepackage{graphicx}
\usepackage{array}
\usepackage{colortbl}
\usepackage{xcolor}
\usepackage{makecell}
\usepackage{float} 

\newcommand*{\menlo}{\fontfamily{lmtt}\fontsize{9}{9}\selectfont }

\newcommand{\myparagraph}[1]{\vspace{1pt} \noindent \textbf{#1}}

\newcommand{\latexblend}{\textsc{LaTexBlend}\xspace}

\newcommand{\refsec}[1]{Section~\ref{sec:#1}}

\newcommand{\reftbl}[1]{Table~\ref{tbl:#1}}

\newcommand{\lblsec}[1]{\label{sec:#1}}

\newcommand{\ignorethis}[1]{}

\definecolor{cvprblue}{rgb}{0.21,0.49,0.74}
\usepackage[pagebackref,breaklinks,colorlinks,allcolors=cvprblue]{hyperref}

\def\paperID{6298} 
\def\confName{CVPR}
\def\confYear{2025}

\title{\latexblend: Scaling Multi-concept Customized Generation \\
with Latent Textual Blending}

\author{
Jian Jin\textsuperscript{\rm 1},
Zhenbo Yu\textsuperscript{\rm 1},
Yang Shen\textsuperscript{\rm 1},
Zhenyong Fu\textsuperscript{\rm 1*\footnotemark[0]},
Jian Yang\textsuperscript{\rm 1*}}

\begin{document}

\twocolumn[{%
\maketitle
\centering
\vspace{-0.3cm}
\noindent
\normalsize\textsuperscript{\rm 1}PCA Lab$^{\dagger}$, Nanjing University of Science and Technology, China \\
{\tt\small \{jinj, zhenboyu, shenyang\_98, z.fu, csjyang\}@njust.edu.cn}


\begin{figure}[H]
\hsize=\textwidth 
\centering
\includegraphics[width=0.99\textwidth]{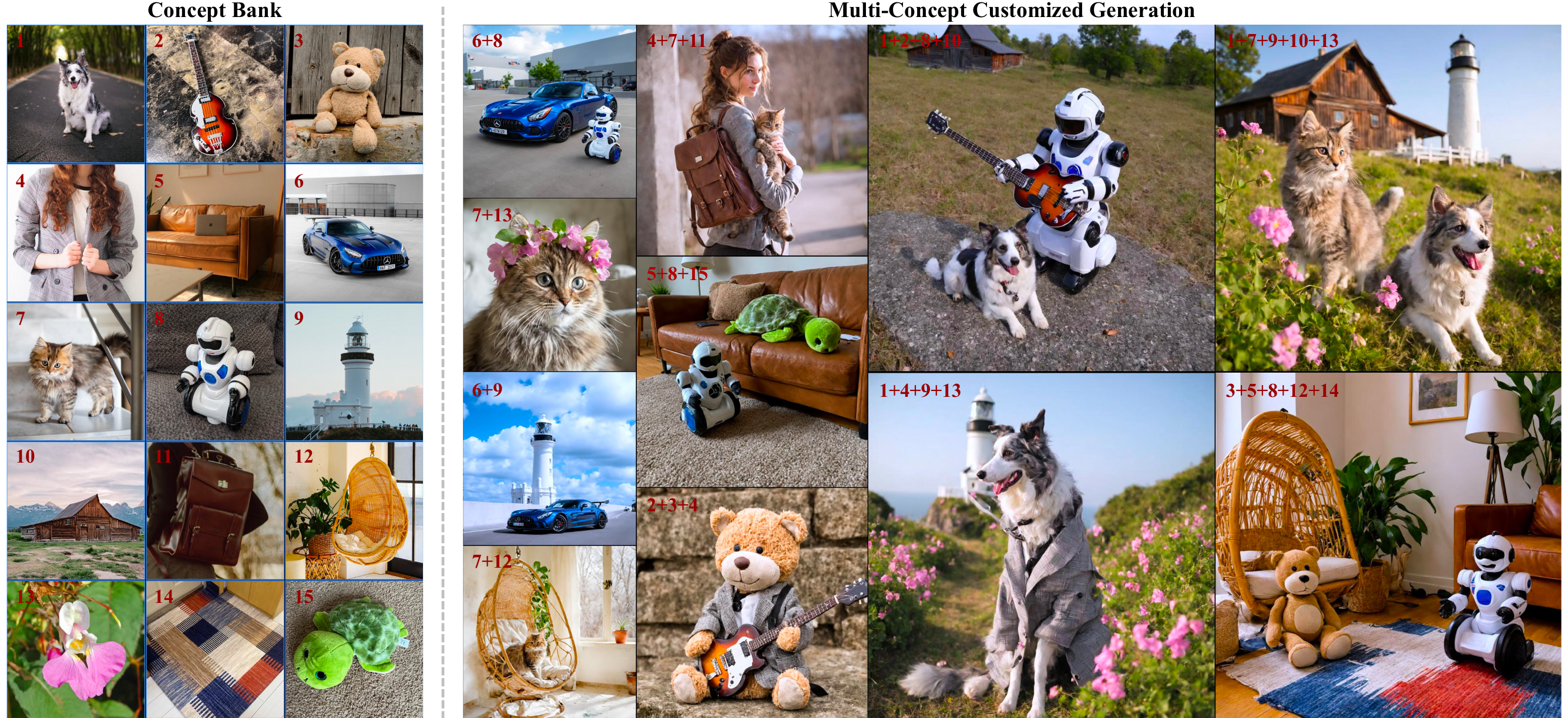}
\caption{\latexblend customizes each personal subject individually and stores it in a concept bank using a compact representation.
At inference, multiple concepts from the bank can be combined seamlessly for multi-concept customized generation without requiring additional tuning. 
\latexblend simultaneously addresses two key challenges in scaling multi-concept generation: ensuring high \textbf{generation quality} (including \textbf{concept fidelity} and \textbf{layout coherence}) and maintaining \textbf{computational efficiency}.
}
    \label{fig:teaser}
\end{figure}
}]

\renewcommand{\thefootnote}{\fnsymbol{footnote}}
\footnotetext[1]{Corresponding authors.}
\footnotetext[2]{
PCA Lab, Key Lab of Intelligent Perception and Systems for High-Dimensional Information of Ministry of Education, and Jiangsu Key Lab of Image and Video Understanding for Social Security, School of Computer Science and Engineering, Nanjing University of Sci. \& Tech.
}
\renewcommand{\thefootnote}{\arabic{footnote}}

\maketitle
\begin{abstract}
Customized text-to-image generation renders user-specified concepts into novel contexts based on textual prompts.
Scaling the number of concepts in customized generation meets a broader demand for user creation, whereas existing methods face challenges with generation quality and computational efficiency.
In this paper, we propose \latexblend, a novel framework for effectively and efficiently scaling multi-concept customized generation.
The core idea of \latexblend is to represent single concepts and blend multiple concepts within a \underline{\textbf{La}}tent \underline{\textbf{Tex}}tual space, which is positioned after the text encoder and a linear projection.
\latexblend customizes each concept individually, storing them in a concept bank with a compact representation of latent textual features that captures sufficient concept information to ensure high fidelity.
At inference, concepts from the bank can be freely and seamlessly combined in the latent textual space, offering two key merits for multi-concept generation: 1) excellent scalability, and 2) significant reduction of denoising deviation, preserving coherent layouts.
Extensive experiments demonstrate that \latexblend can flexibly integrate multiple customized concepts with harmonious structures and high subject fidelity, substantially outperforming baselines in both generation quality and computational efficiency.
\href{https://jinjianrick.github.io/latexblend/}{Project Page}.

\end{abstract}    
\vspace{-5pt}
\section{Introduction}
\label{sec:intro}

Customized text-to-image generation has drawn growing attention for its practicality in personalized creation.
Model customization fine-tunes a pre-trained text-to-image model to implant personalized concepts, providing a promising approach to achieving compelling customized generation.
Recently, customized generation has progressed from single-concept to multi-concept generation, which involves integrating multiple customized subjects into a novel scene based on textual prompts.
Scaling the number of concepts in customized generation will provide greater flexibility and personalization for user creation.

Recent studies have proposed various approaches to advance multi-concept generation.
Some works enhance fine-tuning through multi-concept joint training~\cite{kumari2023multi, liu2023cones} or data augmentation~\cite{han2023svdiff, jang2024identity}, but they face exponential growth in fine-tuning complexity as the number of concepts scales.
Others~\cite{kwon2024concept, kong2024omg} merge single-concept diffusion branches for multi-concept inference, which significantly raises computational overhead.
While a few scalable approaches exist~\cite{liu2023cones2, kong2024omg}, they often compromise on concept fidelity and layout coherence.
Overall, current methods either lack computational efficiency or fall short in generation quality, leaving substantial challenges in scaling customized generation.

\begin{figure}[!t]
 \centering
 \includegraphics[width=0.98\linewidth]{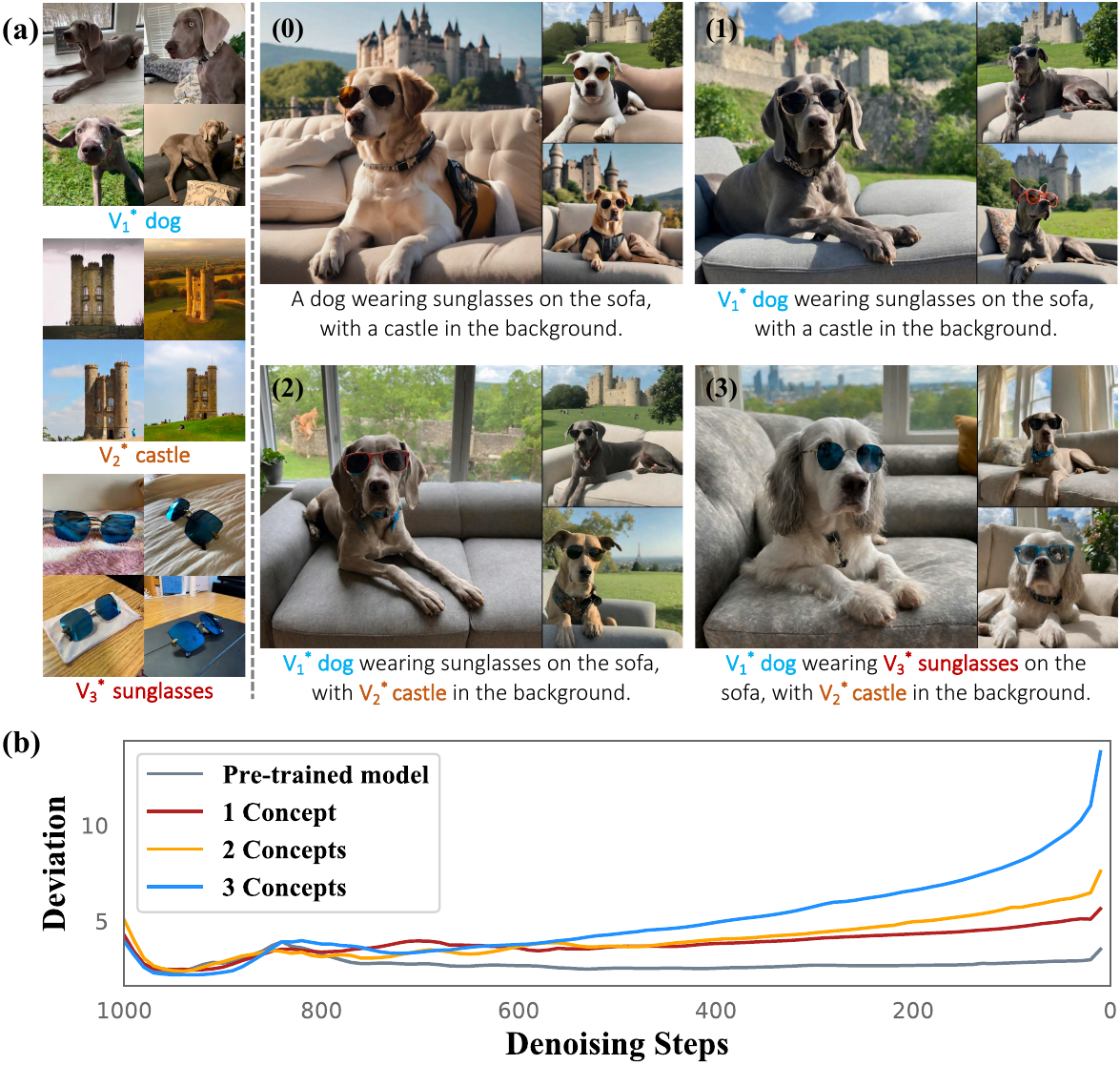}
 \caption{\textbf{Structure degradation and denoising deviation in customized generation.}
 \textbf{(a)}: The images in (0) are generated by the pre-trained model (with 3 different initializations), which are highly aligned with the prompt; images (1)–(3) are generated by Customized Diffusion~\cite{kumari2023multi}.
 In (1)-(3) of Fig. (a), we progressively add the customized concepts - {\menlo "$\text{V}_1^*$ dog"}, {\menlo "$\text{V}_2^*$ castle"}, and {\menlo "$\text{V}_3^*$ sunglasses"} - to the generation, using the same initializations as in (0).
 \textbf{(b)}: We use the magnitude proposed in~\cite{wen2024detecting} to reflect the deviation in image structure.
 Customized generation deviates from the normal denoising process of the pre-trained model, resulting in degraded and memorized layouts (typically single-object-centric).
 This issue worsens as the number of concepts increases.
 }   
 \label{fig:seq_ori}
\end{figure}

We explore the causes of quality degradation in multi-concept generation.
In the normal denoising process, the final structure of generated images is closely tied to the initial noise~\cite{wen2024detecting, wen2023tree}.
The images generated from the same noise initialization have a similar layout.
However, customized generation suffers from denoising deviation regardless of its initialization, induced by the inherent limitations of the customization process.
Conventional customization uses only a few reference images of the subject ($3 \sim 5$), which are typically single-object-centric and lack context diversity.
Therefore, customized concepts struggle to achieve generative generalization on par with general concepts in the pre-trained model.
Generation involving customized concepts deviates abnormally from the normal denoising process, resulting in degraded and memorized image structures, as shown in Fig.~\ref{fig:seq_ori} and Fig.~\ref{fig:compare1}.
More seriously, due to spatial competition and the low co-occurrence probability between customized concepts, the denoising deviation becomes more pronounced as the number of concepts increases.
Therefore, multi-concept generation suffers from severe performance degradation, raising issues like concept omission, concept blending, and concept distortion.

In this paper, we propose \latexblend, a novel framework that efficiently and effectively scales multi-concept customized generation.
The key idea of \latexblend is to represent single concepts and blend multiple concepts in \emph{a \underline{\textbf{La}}tent \underline{\textbf{Tex}}tual space}, which is located after the text encoder and a linear projection.
At fine-tuning, \latexblend customizes each concept individually and stores them in a concept bank as a compact form of \emph{latent textual features}.
Specifically, we introduce an auxiliary textual encoding flow to rectify gradient propagation, consolidating concept information into a compact representation. 
At inference, concepts from the bank can be freely combined \emph{in the latent textual space} for multi-concept customized generation.
We identify that the latent textual space is a pivotal point in conditional diffusion models for multi-concept integration, as it encompasses sufficient customized information without being too deep to induce costly merging.
Besides, blending customized concepts in this space can eliminate their interference in the earlier textual encoding process, thereby reducing denoising deviation.
Therefore, \latexblend can efficiently integrate multiple customized concepts with high subject fidelity and coherent layouts (Fig.~\ref{fig:teaser}).

We conduct extensive experiments to evaluate the proposed \latexblend.
\latexblend simultaneously addresses two key challenges in scaling multi-concept generation, \emph{i.e.}, generation quality and computational efficiency.
Qualitative and quantitative results demonstrate that \latexblend seamlessly integrates multiple customized concepts into the query context with high subject fidelity, while maintaining a coherent layout.
Moreover, \latexblend exhibits excellent scalability in computational efficiency.
The fine-tuning complexity of \latexblend \emph{increases linearly} with the number of concepts, and no additional inference cost incurred as the number of concepts grows.

\section{Related Work}
\label{sec:related_work}

\myparagraph{Text-to-Image Generation.}
Text-to-image (T2I) generation aims to synthesize visually realistic images based on textual descriptions.
Early studies~\cite{ref9, ref10, ref11, ref12} translate natural language into images by employing generative adversarial networks~\cite{ref13}.
Autoregressive models such as DALL-E~\cite{ref15}, Cogview~\cite{ref16}, NUWA~\cite{ref17}, and Parti~\cite{ref18}, reframe text-to-image generation as a sequence-to-sequence problem.
Recent advances~\cite{ref19, ref20, ref4, ref21, podellsdxl, esser2024scaling} in text-to-image generation harness diffusion models (DMs)~\cite{ref22, ho2020denoising, ref43} as the generative backbone. These models generate images through a denoising task while incorporating text conditions during the denoising process.

\myparagraph{Single-concept Customized T2I Generation.}
Single-concept customized T2I generation aims to incorporate a new concept into pre-trained T2I models, generating images of this concept in new contexts.
Optimization-based methods fine-tune pre-trained T2I models for customized generation, implanting the customized concept by inverting it into a word embedding vector~\cite{ref6, voynov2023p, alaluf2023neural} or bounding it with special tokens and model parameters~\cite{ref2, tewel2023key}.
Subsequent works focus on developing more efficient methods~\cite{kumari2023multi, gal2023designing, han2023svdiff} or on extending to image editing~\cite{zhang2023sine, jin2025unicanvas}.
Another line research on customized generation is learning-based method~\cite{wei2023elite, ding2024freecustom, zhang2024ssr, wang2024ms}, which aims to train unified models capable of personalizing diverse subject inputs.
The proposed \latexblend is an optimization-based method.

\myparagraph{Multi-concept Customized Generation.}
A more challenging task is multi-concept customized generation, which entails integrating multiple customized concepts into a single image.
Some works enhance fine-tuning through multi-concept joint training~\cite{kumari2023multi, liu2023cones}, data augmentation~\cite{han2023svdiff, jang2024identity}, or by merging multiple single-concept diffusion pipes~\cite{jiang2024mc, kong2024omg}, which lack computational efficiency.
Others~\cite{liu2023cones2} are more efficient but fall short in generation quality.
Our method addresses both generation quality and computational efficiency in scaling multi-concept generation.

\section{Method}\label{method}

\begin{figure*}[!t]
 \centering
 \includegraphics[width=0.98\textwidth]{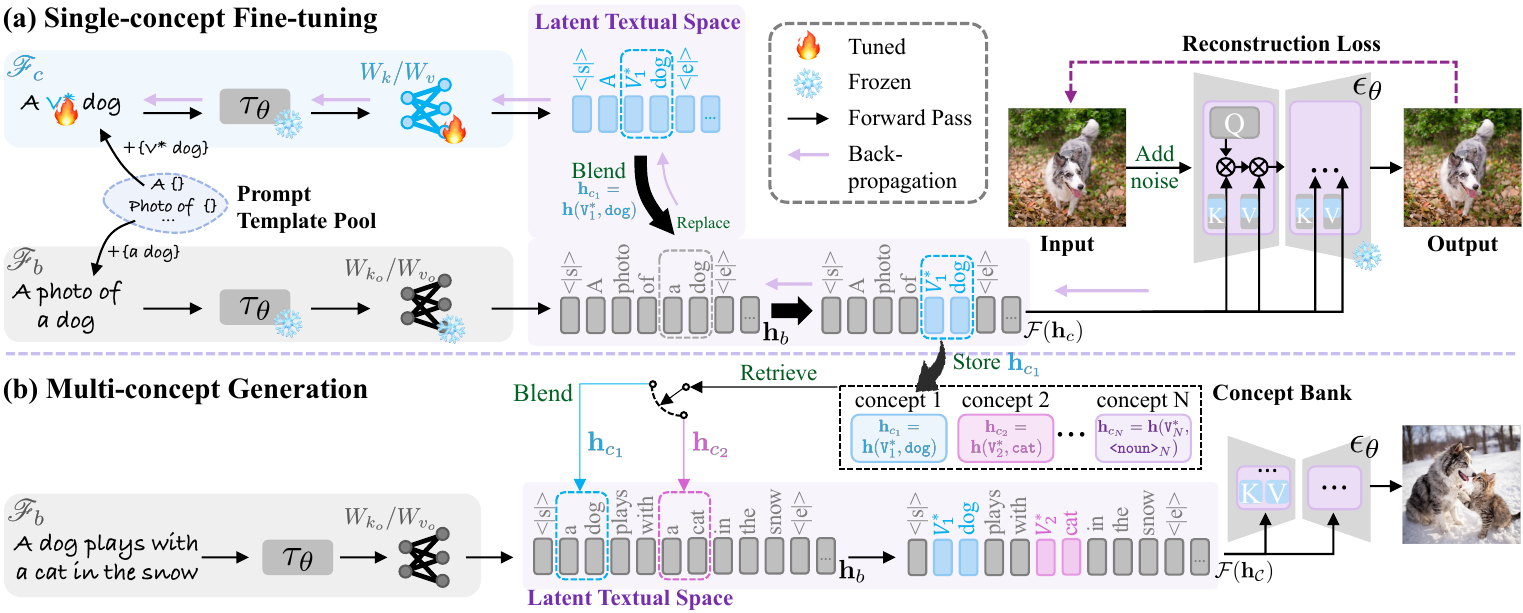}
 \vspace{-5pt}
 \caption{\textbf{Overall framework of the proposed \latexblend.}  
 \textbf{(a)} \latexblend customizes each concept individually and stores them in a concept bank with a compact representation of latent textual features.
 \textbf{(b)} At inference, concepts from the bank can be seamlessly combined in the latent textual space on the fly for multi-concept generation, without needing any additional tuning.
 }
 \vspace{-10pt}
 \label{fig:model_merge}
\end{figure*}

\latexblend customizes each concept individually and stores them in the concept bank with a novel representation of \emph{latent textual features} (Section~\ref{single_concept}).
At inference, concepts from the bank can be seamlessly combined \emph{in the latent textual space} on the fly for multi-concept generation, without needing any additional tuning (Section~\ref{multi_concept}).

\subsection{Preliminary of Customized Generation}
Text-to-image (T2I) model customization binds a custom concept $c$ with special tokens $\mathcal{T}$ by fine-tuning pre-trained T2I models using $S$ image-prompt pairs, $\mathcal{D}=\left\{(\bm{X}_i, \bm{P}_i(\mathcal{T})) \right\}_{i=1}^S $.
Here, $\bm{X}_i$ is a reference image of $c$, and $\bm{P}_i(\mathcal{T})$ denotes the corresponding textual prompt that contains $\mathcal{T}$.
$\mathcal{T}$ is typically in the form {\menlo "V* <noun>"}, where {\menlo "V*"} is an identifier token and {\menlo "<noun>"} is a coarse class descriptor of the subject (\emph{e.g.} cat, dog).
We use Latent Diffusion Models (LDMs)~\cite{ref4} as the generative backbone.
The customization process is achieved through a T2I reconstruction task using the following squared error loss:
\begin{equation}\label{loss1}
    \mathbb{E}_{z\sim \mathcal{E}(\bm{X}_i), \epsilon \sim \mathcal{N}(0,1),\bm{P}_i,t}\left[w_t \Vert \epsilon - \epsilon_\theta (z_t; \tau_\theta({\bm{P}_i}), t) \Vert_2^2  \right]  \,,
\end{equation}
where $z_t := \alpha_t z + \sigma_t \epsilon$ is a noised latent image derived from the ground-truth image $\bm{X}_i$ at timestep $t$, and $w_t$, $\alpha_t$, $\sigma_t$ are time-dependent terms that determine the loss weight and noise schedule; $\epsilon_\theta$ is a denoising autoencoder; $\mathcal{E}$ and $\tau_\theta$ are an image encoder and a text encoder.
At inference, given the query prompt $\bm{P}(c)$, we can generate the target image from noise $\mathbf{z}_T\sim\mathcal{N}(0,1)$ through iterative updates:
\begin{equation}\label{inference1}
    z_{t-1} = \texttt{update} (z_t, \hat{\epsilon}_t, t, t-1, \epsilon_{t-1}) \,,
\end{equation}
where $\hat{\epsilon}_t=\epsilon_\theta(z_t; \tau_\theta(\bm{P}(c)), t)$, and $\texttt{update}$ represents sampling methods like DDPM~\cite{ho2020denoising} and DDIM~\cite{ref43}.

\subsection{Single-concept Fine-tuning}\label{blending in latent textual space}\label{single_concept}

In T2I generation, the query prompt $\bm{P}$ is first encoded into a textual feature $\mathbf{e}=\tau_\theta({\bm{P}})\in \mathbb{R}^{M\times d_t}$ by a text encoder $\tau_\theta$, where $M$ is a fixed text sequence length and $d_t$ is the feature dimension.  
Two projection matrices, $W_k$ and $W_v \in \mathbb{R}^{d_t\times d_l}$, are then used to map $\mathbf{e}$ to latent textual features:
\begin{equation}
\begin{split}
    \mathbf{K} = \texttt{proj}_k(\mathbf{e}) = W_k\mathbf{e} \,, \\
    \mathbf{V} = \texttt{proj}_v(\mathbf{e}) = W_v\mathbf{e} \,,
\end{split}
\end{equation}
where $\mathbf{K}, \mathbf{V} \in \mathbb{R}^{M\times d_l}$, and $d_l$ is the dimension of latent features.
$\mathbf{K}$ and $\mathbf{V}$ serve as the key and value features for the cross-attention blocks, integrating textual conditions into the diffusion model.
We refer to the process of converting a textual prompt $\bm{P}$ into latent textual features $\mathbf{h}=\{\mathbf{K}, \mathbf{V}\}$ as the textual encoding flow:
\begin{equation}
    \mathscr{F}: \bm{P} \xrightarrow{\texttt{proj}(\tau_\theta(\cdot))}\mathbf{h}\,.
\end{equation}
Custom diffusion~\cite{kumari2023multi} has demonstrated that fine-tuning only the embedding of identifier token {\menlo "V*"} along with the projection matrices $W_k$ and $W_v$ is adequate for concept implantation. 
Since all these parameters are contained within $\mathscr{F}$, we can deduce that the output of $\mathscr{F}$ (\emph{i.e.}, $\mathbf{h}$) already encompasses sufficient concept-related information.

In $\mathscr{F}$, all tokens attend to {\menlo "V*"} in $\tau_\theta$ and share the customized projection matrices. Therefore, the concept-related information is dispersed across all $M$ token features of $\mathbf{h}$.
To disentangle a compact concept representation $\mathbf{h}_c$ from the entire $\mathbf{h}$, we propose a novel single-concept fine-tuning strategy that consolidates concept-related information into the latent textual features of concept-relevant tokens $\mathcal{T}$ (\emph{i.e.}, {\menlo "V*"} and {\menlo "<noun>"}), as shown in Fig.~\ref{fig:model_merge} (a).
Specifically, apart from the regular textual encoding flow, denoted as $\mathscr{F}_c$, we introduce an additional base flow $\mathscr{F}_b$ that shares the same text encoder with $\mathscr{F}_c$ but uses frozen pre-trained projection matrices $\{W_{k_o}, W_{v_o}\}$.
The prompt of $\mathscr{F}_b$ is the same as that of $\mathscr{F}_c$, except the learnable identifier {\menlo "V*"} replaced by frozen common articles (\emph{e.g.}, ``a", ``the").
We denote the output of $\mathscr{F}_b$ as $\mathbf{h}_b$, while $\mathbf{h}_c$ is the latent textual features of $\mathcal{T}$ in the output of $\mathscr{F}_c$.
During fine-tuning, $\mathbf{h}_c$ is blended with $\mathbf{h}_b$ in the \emph{latent textual space}:
\begin{equation}\label{tuning_blend}
    \mathcal{F}(\mathbf{h}_c) = \texttt{Blend} (\mathbf{h}_b; \mathbf{h}_c)\,,
\end{equation}
where in the $\texttt{Blend}()$ operation, $\mathbf{h}_c$ replaces the latent textual features of the corresponding tokens (\emph{i.e.}, {\menlo "<noun>"} and its article) in $\mathbf{h}_b$.
The blended latent textual feature $\mathcal{F}(\mathbf{h}_c)$ is used as the new key and value features for attention calculation, and the fine-tuning loss in Eq.~\ref{loss1} is modified as:
\begin{equation}
    \mathbb{E}_{z\sim \mathcal{E}(x), \epsilon \sim \mathcal{N}(0,1),L^b,t}\left[w_t \Vert \epsilon - \epsilon_\theta (z_t, \mathcal{F}(\mathbf{h}_c), t) \Vert_2^2  \right]  \,.
\end{equation}
With this modification, the gradient can only backpropagate through $\mathbf{h}_c$ to the learnable branch $\mathscr{F}_c$ for parameter optimization.
In the forward pass, concept-related information derived from the tuned parameters is passed only to $\mathbf{h}_c$ within the entire conditioning $\mathcal{F}(\mathbf{h}_c)$. 
This fine-tuning process aims to accurately reconstruct reference images using only concept-related information in $\mathbf{h}_c$.
Therefore, concept-related information is forced to concentrate in $\mathbf{h}_c$ to sufficiently represent the target concept $c$.
We store $\mathbf{h}_c$ in a concept bank for subsequent multi-concept generation.

\myparagraph{Position Invariance.} At inference, the blending position of $\mathbf{h}_c$ can vary anywhere in the prompt. Therefore, $\mathbf{h}_c$ must be position-invariant to ensure correct functionality regardless of its blending position.
To address this issue, we create a template pool containing 7 different prompt templates (\emph{e.g.}, {\menlo "A \{\}.", "Photo of \{\}."}) of varying lengths.
During fine-tuning, we randomly draw different templates from the pool to construct the prompts of $\mathscr{F}_b$ and $\mathscr{F}_c$.
By varying the extraction and insertion positions of $\mathbf{h}_c$, this prompt variability strategy mitigates positional dependency and facilitates position invariance for $\mathbf{h}_c$.

\begin{figure}[!t]
 \centering
 \includegraphics[width=0.98\linewidth]{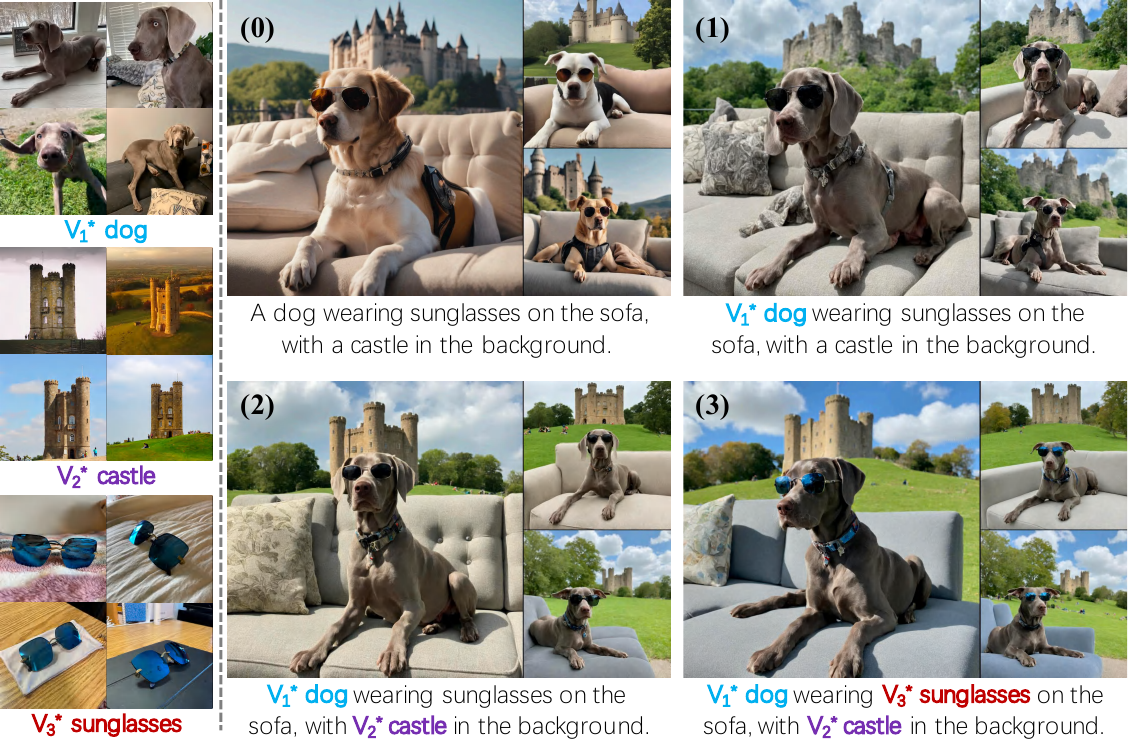}
 \vspace{-5pt}
 \caption{\textbf{Mitigation of image structure degradation.}
 The images in (0) are generated by the pre-trained model with 3 different initializations.
 In images (1)-(3), we progressively blend customized concepts - {\menlo "$\text{V}_1^*$ dog"}, {\menlo "$\text{V}_2^*$ castle"}, and {\menlo "$\text{V}_3^*$ sunglasses"} - into the context of image (0).
 \latexblend effectively mitigates structure degradation caused by customized concepts, blending high-fidelity subject appearances while maintaining coherent layouts.
 }
 \vspace{-10pt}
 \label{fig:seq_blend}
\end{figure}

\subsection{Multi-concept Generation}\label{multi_concept}
We blend single concepts in the \emph{latent textual space} for multi-concept generation, as shown in Fig.~\ref{fig:model_merge} (b).
For a multi-concept generation task involving $N$ customized concepts $\mathcal{C} = \{c_1, c_2, \cdots, c_N \}$, we retrieve the corresponding latent textual features $\mathbf{h}_\mathcal{C} = \{ \mathbf{h}_{c_1}, \mathbf{h}_{c_2}, \cdots, \mathbf{h}_{c_N} \}$ from the concept bank.
We first create a base textual encoding flow $\mathscr{F}_b$ that uses the vanilla pre-trained projection matrices $\{W_{k_o}, W_{v_o}\}$.
For a query prompt $\bm{P}(\mathcal{C})$, $\mathscr{F}_b$ outputs the base latent textual features $\mathbf{h}_b$.
Since the latent features in $\mathbf{h}_\mathcal{C}$ sufficiently represent the customized concepts and possess position invariance, 
we integrate concepts into the query context by selectively blending $\mathbf{h}_\mathcal{C}$ with $\mathbf{h}_b$ using simple latent feature replacements:
\begin{equation}\label{multi_merge}
    \mathcal{F}(\mathbf{h}_\mathcal{C}) = \texttt{Blend}(\mathbf{h}_b; \mathbf{h}_{c_1}, \mathbf{h}_{c_2}, \cdots, \mathbf{h}_{c_N})\,,
\end{equation}
where the $\texttt{Blend}()$ operation is analogous to that in Eq.~\ref{tuning_blend} but extended to the multi-concept case.
$\mathcal{F}(\mathbf{h}_\mathcal{C})$ then serves as the key and value features for multi-concept generation.

\myparagraph{Context Generalization.}
A remarkable property of the concept representation in the latent textual space is that it is derived from a simple textual template yet demonstrates robust context generalization.
We show examples in Fig.~\ref{fig:seq_blend},
\latexblend can inject customized appearances while greatly preserving the overall coherent layout arranged by the pre-trained model.
We infer that blending concepts in the latent textual space eliminates the interference of customized tokens with other tokens in the earlier textual encoding process (\emph{i.e.}, within $\mathscr{F}$), thereby reducing deviations from the normal denoising trajectory.
This allows \latexblend to seamlessly integrate customized concepts into target contexts in a coherent manner. 

\myparagraph{Blending Guidance.}
\latexblend can greatly mitigate the degradation of the image structure arranged by the pre-trained model for customized generation.
However, pre-trained T2I models may also struggle with challenging cases~\cite{bakr2023hrs, gani2023llm}, such as generating similar subjects or a large number of subjects together. 
In these cases, diffusion models produce incorrect cross-attention maps $\mathcal{A}$, which directly determine the spatial layout and structure of the generated image~\cite{ref42}.
Guidance~\cite{ho2022classifier, dhariwal2021diffusion, jiang2024mc} can adjust the sampling process toward specific targets by modifying the update rule in Eq.~\ref{inference1}.
Inspired by works~\cite{rassin2024linguistic, epstein2023diffusion}, we propose \emph{blending guidance} to steer the inference process toward the desired attention rectification, thereby boosting multi-concept generation.
For the query prompt $\bm{P}(\mathcal{C})$, the tokens to be considered are $\mathcal{T} = \{\mathcal{T}_1, \cdots, \mathcal{T}_N\} = \{\{\text{{\menlo V}}_1^*, \text{{\menlo <noun>}}_1 \}, \cdots, \{ \text{{\menlo V}}_N^*, \text{{\menlo <noun>}}_N \}\}$.
We denote the attention map of token $t$ as $\mathcal{A}_t$, and denote the spatial overlap between two attention maps $\mathcal{A}_{t_1}$ and $\mathcal{A}_{t_2}$ as $\textbf{O}(\mathcal{A}_{t_1}, \mathcal{A}_{t_1})$.
The blending guidance consists of two terms.
The first term encourages maximizing the overlap between the attention map of all identifier tokens {\menlo "V*"} and their class descriptors {\menlo "<noun>"} to enhance identity binding:
\begin{equation}\label{g1}
    g_1 = -\sum_{i=1}^N \textbf{O}(\mathcal{A}_{\text{{\menlo V}}_i^*}, \mathcal{A}_{\text{{\menlo <noun>}}_i}) \,.
\end{equation} 
The second term aims to pull apart the functional regions of different concept tokens and minimize the overlap, thereby preventing identity mixing between customized subjects:
\begin{equation}\label{g2}
    g_2 = \frac{1}{2N} \sum_{i=1}^N \sum_{k \in \mathcal{T} \backslash \{ \mathcal{T}_i \}} \textbf{O}(\mathcal{A}_{\text{{\menlo V}}_i^*}, \mathcal{A}_k) + \textbf{O}(\mathcal{A}_k, \mathcal{A}_{\text{{\menlo <noun>}}_i}) \,.
\end{equation}
The blending guidance is incorporated into the score estimate, modifying the update direction $\hat{\epsilon}_t$ in Eq.~\ref{inference1} as:
\begin{equation}\label{total_loss}
    \hat{\epsilon}^{\prime}_t = \hat{\epsilon}_t + \lambda \nabla_{z_t} (g_1 + g_2) \,.
\end{equation}
The obtained $\hat{\epsilon}^{\prime}_t$ replaces $\hat{\epsilon}_t$ in Eq.~\ref{inference1} to update the latents. The iterates are nudged down the gradient of the guidance, steering the sampling process toward the desired output.

\myparagraph{Customized Generation with Layout Conditions.}
The proposed \latexblend also paves the way for efficiently and effectively scaling the number of concepts in customized layout-to-image generation.
By combining with existing layout-to-image diffusion models~\cite{kim2023dense}, \latexblend can integrate multiple customized subjects into spatial arrangements that conform to the provided layouts.

\begin{figure}[!t]
    \centering
    \includegraphics[width=\linewidth]{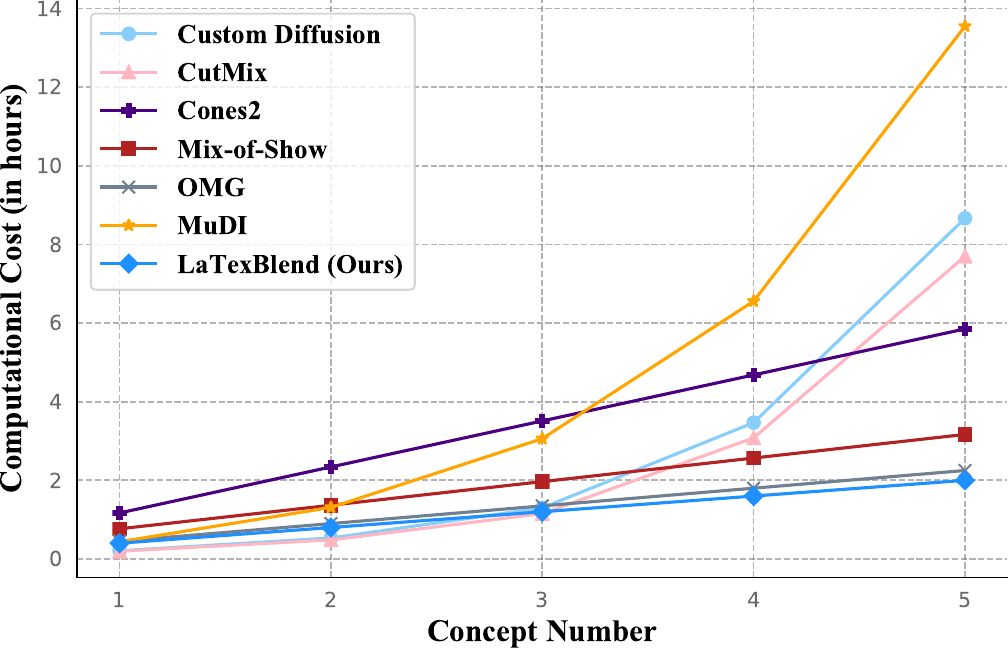}
    \vspace{-15pt}
    \caption{\textbf{Comparison of fine-tuning costs.}
    The fine-tuning cost of \latexblend increases linearly as the number of concepts grows.
    Mix-of-Show~\cite{gu2024mix} requires extra tuning for different concept combinations.
    Although OMG~\cite{kong2024omg} is also efficient for fine-tuning, 
    its inference-time cost is high.
    }
    \label{fig:cost}
    \vspace{-15pt}
\end{figure}

\section{Experiments}
\label{sec:exper}

\begin{figure*}[!t]
 \centering
 \includegraphics[width=0.98\textwidth]{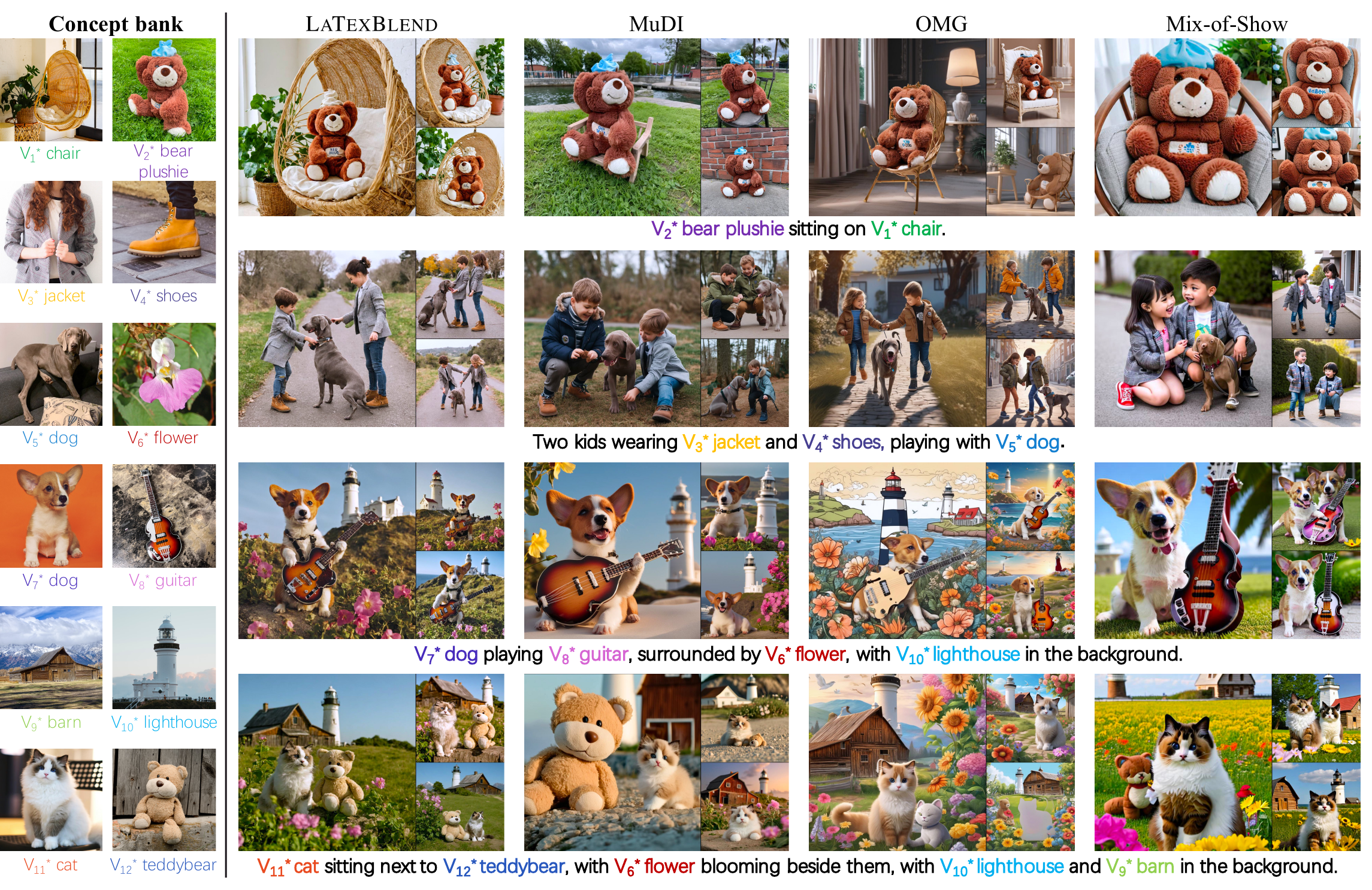}
 \vspace{-8pt}
 \caption{\textbf{Visual comparison with baselines.} 
 We compare \latexblend with recent, competitive baselines. The baselines exhibit significant subject omission and degradation in concept fidelity.
 In contrast, \latexblend renders multiple subjects to the query context with high fidelity and coherent layouts, while also capturing complex semantics such as inter-subject interactions and actions.
 }
 \label{fig:compare1}
 \vspace{-10pt}
\end{figure*}

\begin{figure}[!t]
 \centering
 \includegraphics[width=0.98\linewidth]{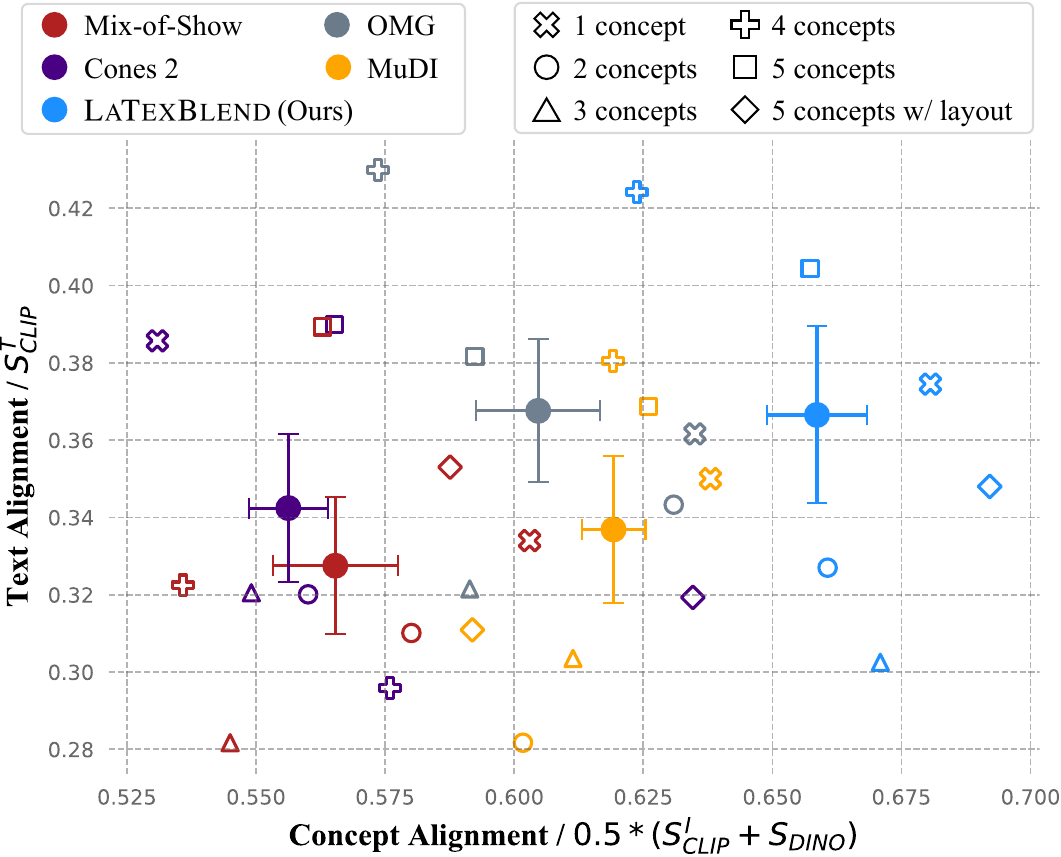}
 \vspace{-5pt}
 \caption{\textbf{Quantitative results.} 
 Marker shape indicates the number of concepts, and color represents the method used. Solid circles denote aggregated results. The $x$-axis represents the average of $S_{\text{CLIP}}^\mathrm{I}$ and $S_{\text{DINO}}$. Cones 2 heavily relies on the layout condition. Our method significantly outperforms baselines. Detailed data are provided in the supplementary material.
 }   
 \label{fig:quantitative}
 \vspace{-20pt}
\end{figure}

\subsection{Experimental Setups}
\myparagraph{Datasets.}
For a fair comparison,  we conduct experiments on $30$ subjects from previous studies~\cite{ref6, ref2, kumari2023multi}.
These subjects cover various categories, such as buildings, pets, and objects, each depicted by several reference images.
More details are provided in the supplementary material.

\myparagraph{Evaluation Metrics.} 
Following previous studies, we evaluate our method in two aspects:
1) Textual alignment, assessed by the average CLIP~\cite{radford2021learning} similarity $(S_{\text{CLIP}}^\mathrm{T})$ between the generated images and their textual prompts.
2) Concept alignment, measured by the CLIP and DINO~\cite{caron2021emerging} similarities ($S_{\text{CLIP}}^\mathrm{I}$ and $S_{\text{DINO}}$) between the generated images and the reference images.
For generations involving multiple subjects, we compute the image similarity between the generated subjects and the corresponding reference images individually and then report the mean value.

\myparagraph{Baselines.}
We compare \latexblend with existing representative multi-concept generation methods: Custom Diffusion~\cite{kumari2023multi}, CutMix~\cite{han2023svdiff}, Cones 2~\cite{liu2023cones2}, Mix-of-Show~\cite{gu2024mix}, OMG~\cite{kong2024omg}
, and MuDI~\cite{jang2024identity}.
In qualitative and quantitative analyses, we focus on more recent and competitive methods~\cite{liu2023cones2, gu2024mix, kong2024omg, jang2024identity} for comparison.
We omit some methods~\cite{liu2023cones, tewel2023key, kwon2024concept} because they lack released code and are outdated.
The implementation details are provided in the supplementary material.

\begin{figure*}[!t]
 \centering
 \includegraphics[width=0.98\textwidth]{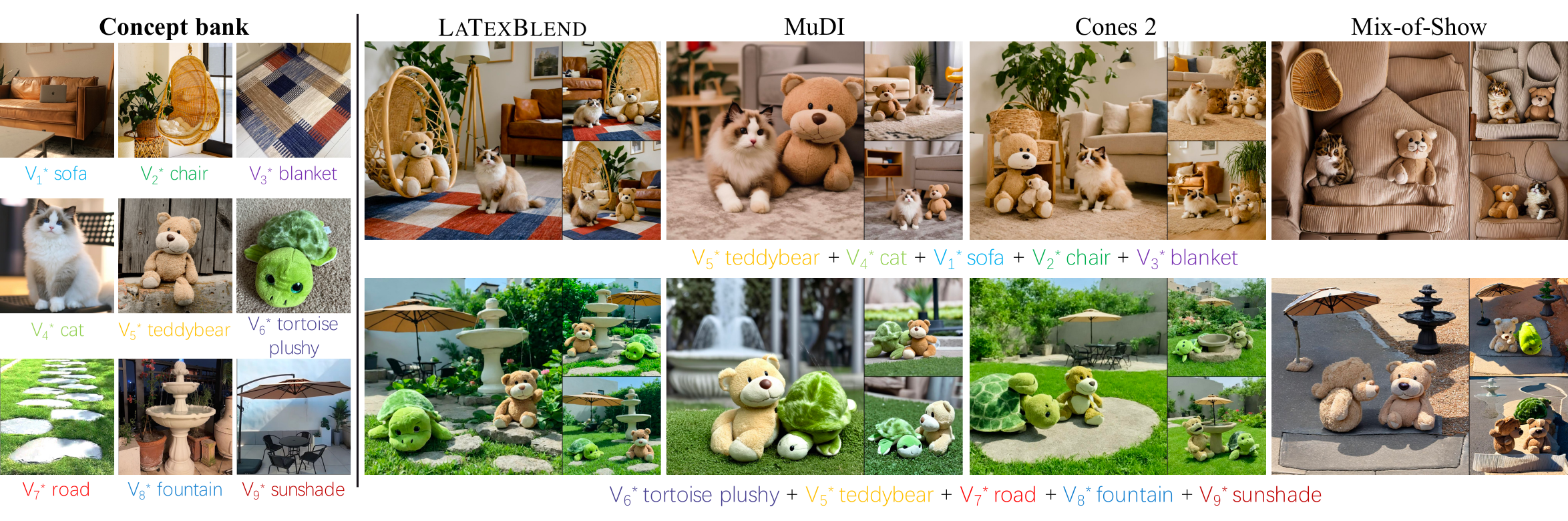}
 \vspace{-8pt}
 \caption{\textbf{Visual comparison of generations with additional layout conditioning.} 
 We compare with methods that support layout conditioning. 
 The baselines typically struggle to generate more than 3 subjects while preserving their fidelity.
 Compared to the baselines, \latexblend demonstrates a higher success rate of subject presence and better subject fidelity, further validating its effectiveness.
 }
 \vspace{-10pt}
 \label{fig:compare2}
\end{figure*}

\begin{table}[!t]
\centering
\setlength{\tabcolsep}{5pt}
\resizebox{\linewidth}{!}{
\begin{tabular}{@{\extracolsep{4pt}}cc c  cc@{} }
\toprule

\textbf{\multirow{2}{*}{Method}}   
& \multirow{2}{*}{\shortstack[c]{Concept\\ Alignment ($\uparrow $) }}
& \multirow{2}{*}{\shortstack[c]{Prompt\\ Alignment ($\uparrow $) } }
& \multirow{2}{*}{\shortstack[c]{Overall\\ Quality ($\uparrow $)} } \\ \\ 

\midrule[0.5pt]
Mix-of-Show~\cite{gu2024mix} & $\text{2.26}$ & $\text{2.55}$ & $\text{2.31}$ \\
Cones 2~\cite{liu2023cones2} & $\text{1.92}$ & $\text{3.26}$ & $\text{3.12}$ \\
OMG~\cite{kong2024omg} & $\text{2.84}$ & $\text{3.66}$ & $\text{1.53}$ \\
MuDI~\cite{jang2024identity} & $\text{3.24}$ & $\text{2.83}$ & $\text{3.54}$ \\

\midrule

\textbf{\latexblend (Ours)} & $\textbf{4.33}$ & $\textbf{4.16}$ & $\textbf{4.76}$ \\

\bottomrule
\vspace{-10pt}
\end{tabular}
}
\vspace{-8pt}
\caption{\textbf{User study.} 
The average ranking scores of the human evaluation range from 0 (worst) to 5 (best).
Our method achieved the highest scores across all three metrics.
}
\label{tbl:user_study}
\vspace{-10pt}
\end{table}

\subsection{Main Results.}

\myparagraph{Comparison of Computational Costs.}
We compare \latexblend with baselines in terms of computational cost as the number of concepts increases.
The first key aspect is the fine-tuning cost, measured by the product of fine-tuning complexity and the time required for each individual customization.
The comparison results are presented in Fig.~\ref{fig:cost}.
Custom Diffusion~\cite{kumari2023multi}, CutMix~\cite{han2023svdiff}, and MuDI~\cite{jang2024identity} require joint training or data augmentation for multi-concept fine-tuning.
These methods train separate models for each subject combination, leading to an exponential increase in fine-tuning complexity as the number of concepts grows.
Mix-of-Show~\cite{gu2024mix} fuses multiple existing concept LoRAs but still requires additional retraining for new subject combinations.
Cones 2~\cite{liu2023cones2} fine-tunes the text encoder for customization, significantly increasing the convergence time.
The fine-tuning complexity of \latexblend increases linearly with the number of concepts, and the model converges quickly for single-concept customization. Once the concept bank is established, no additional tuning costs are required for any subject combination.
Although OMG~\cite{kong2024omg} is also scalable for fine-tuning, its inference-time computation is high and increases proportionally to the number of concepts.
Regarding \latexblend, no additional inference cost is incurred as the number of concepts increases.

\myparagraph{Qualitative Comparison.}
We conduct qualitative comparisons between \latexblend and baseline methods in multi-concept generation. 
For each competing method, we randomly generate 10 images per case and select the best 3 for visual comparison. The results are shown in Fig.~\ref{fig:compare1}.
Baseline generations exhibit issues such as subject omission and fidelity degradation.
MuDI and Mix-of-Show suffer from degradation in images structure, often resulting in single-object-centric subjects.
OMG’s performance heavily depends on an additional segmentation model~\cite{kirillov2023segment} and the quality of initialization.
In contrast, \latexblend can render multiple subjects to the query context with high fidelity and coherent layouts, while also capturing complex semantics, such as inter-subject interactions and actions.
This verifies that concept representation captures sufficient concept information, and blending concepts within the latent textual space significantly mitigates structural degradation in images.
We also compare \latexblend with existing learning-based methods~\cite{ding2024freecustom, zhang2024ssr, wang2024ms} and show clear advantages.
We provide more visual comparisons with Cones 2~\cite{liu2023cones2}, Custom Diffusion~\cite{kumari2023multi}, and learning-based methods in the supplementary materials. 
Furthermore, we perform multi-concept generation with additional layout conditioning, as shown in Fig.~\ref{fig:compare2}.
Compared to the baselines, \latexblend demonstrates a higher success rate of subject presence and better subject fidelity.

\myparagraph{Quantitative Comparison.}
We calculate the text alignment and concept alignment of \latexblend and the baselines with an increasing number of concepts, presenting the results in Fig.~\ref{fig:quantitative}. 
For each method, we randomly generate 10 results per case and report the average scores for comparison. 
\latexblend significantly outperforms previous methods in concept-alignment scores, demonstrating high concept fidelity and powerful concept editability.
This further validates the effectiveness of \latexblend in representing concepts and mitigating structural degradation in customized generation.

\myparagraph{User Study.}
We further evaluate the proposed \latexblend through a user study.
The study includes 20 sets of generation cases with 25 participants. In each case, users are provided with a textual prompt, reference images, and corresponding generations from different methods.
Users are asked to score each generation on a scale from 0 to 5 (with 0 being the worst and 5 the best) based on three criteria: 
1) presence of the target concepts and their alignment with reference images, 2)  alignment with the given prompt, and 3) overall quality and authenticity of the generated images. The average scores are reported in~\reftbl{user_study}.
Our method achieved the highest scores across all 3 metrics, demonstrating high perceptual quality and strong user preference.
This further demonstrates the substantial potential of \latexblend for real-world applications.
We provide more details in the supplementary material.

\subsection{Ablation Studies}

\begin{figure}[!t]
 \centering
 \includegraphics[width=\linewidth]{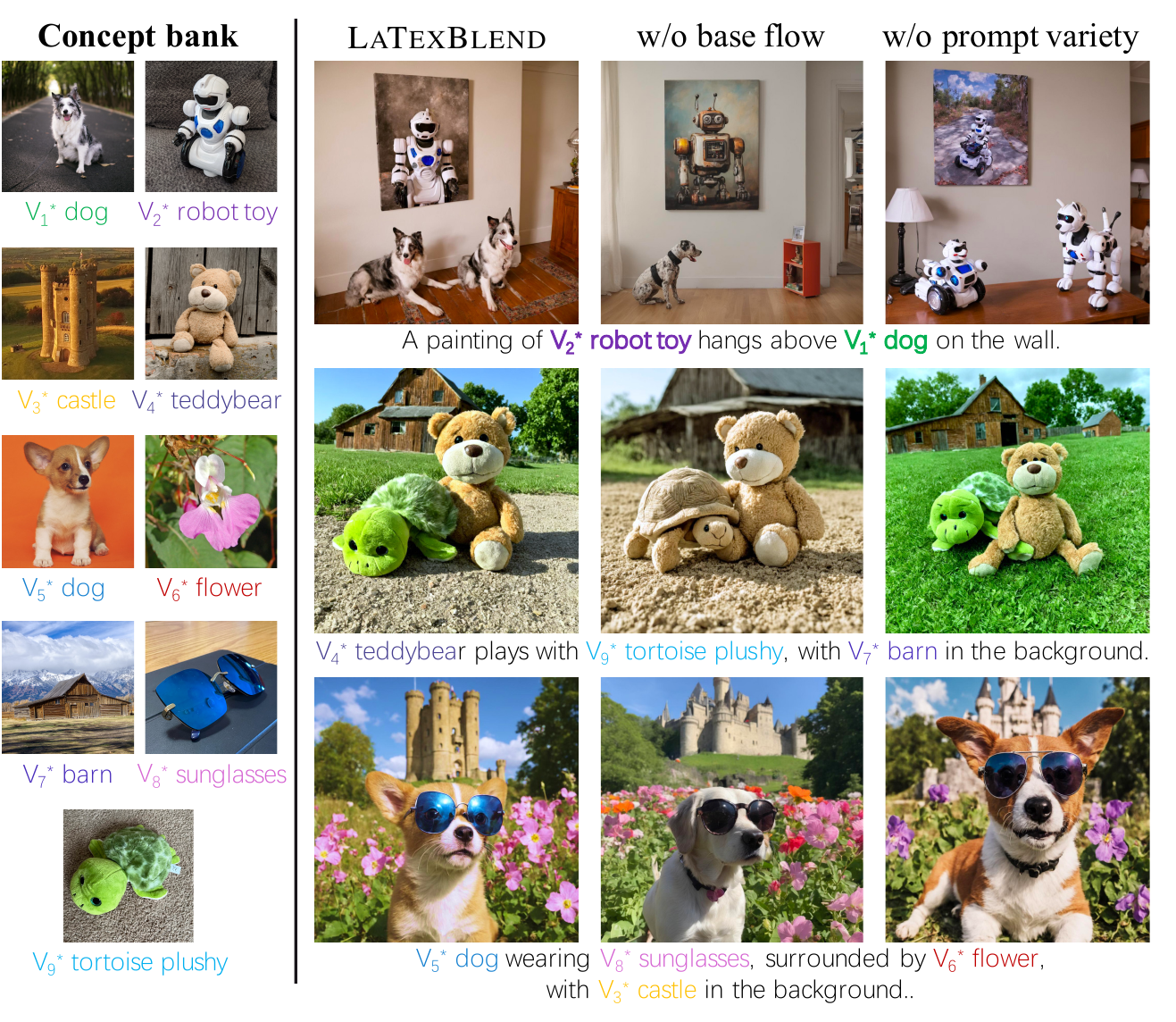}
 \caption{ \textbf{Ablation of two key components in single-concept fine-tuning.}
 Without the base flow, \latexblend would suffer significant degradation in concept fidelity. Using a fixed prompt template results in artifacts in the blended images.
 }   
 \vspace{-10pt}
 \label{fig:ablation1}
\end{figure}

\myparagraph{Key Components in Single-concept Customization.}
We ablate the following two components in single-concept customization:
1) We introduce a frozen base embedding flow, $\mathscr{F}_b$, to rectify information dispersion in latent textual features. In the \emph{``w/o base flow"} scenario, we remove the base flow and fine-tune using only the single learnable concept flow $\mathscr{F}_c$.
2) We create a prompt template pool with templates of varying lengths. In the \emph{``w/o prompt variety"} scenario, we use fixed prompt templates for both $\mathscr{F}_b$ and $\mathscr{F}_c$.
We present quantitative results in~\reftbl{ablation} and show sample generations in Fig.~\ref{fig:ablation1}.
As demonstrated, the base flow plays a crucial role in concentrating concept-related information into the target tokens.
Without it, \latexblend would collapse, resulting in significant degradation of concept fidelity.
The variety in prompt templates is also essential for the proper functioning of latent textual features.
When using a fixed prompt template for fine-tuning, the generated images often exhibit artifacts such as perspective deformation, layout distortion, and subject mixing.

\begin{figure}[!]
 \centering
 \includegraphics[width=0.98\linewidth]{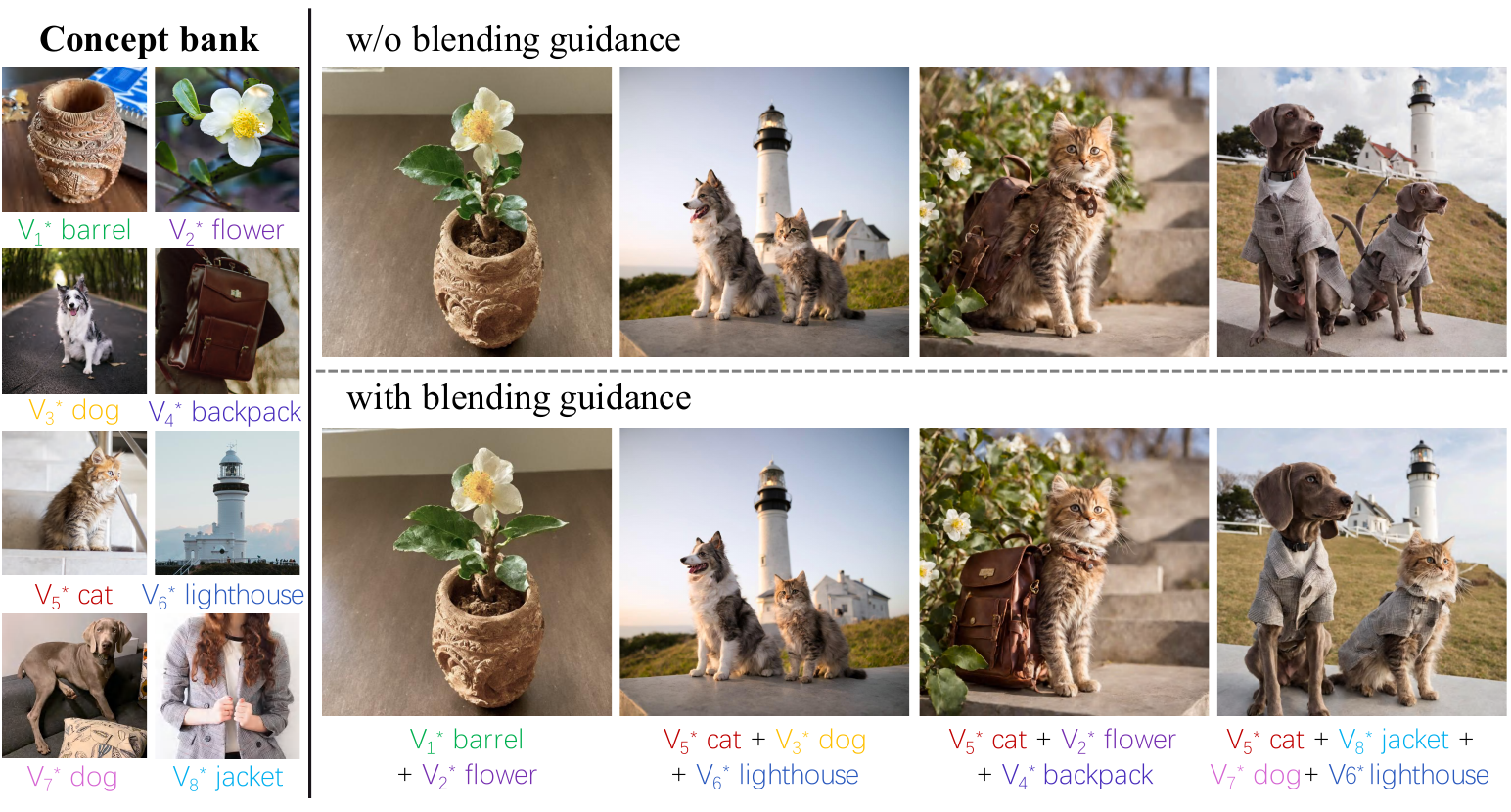}
 \vspace{-5pt}
 \caption{\textbf{Ablation on blending guidance.} Blending guidance
 offers a subtle improvement with fewer concepts, while enhancing multi-concept generation as the number of concepts increases.
 }   
 \label{fig:ablation2}
 \vspace{-5pt}
\end{figure}

\begin{table}[!t]
\centering
\setlength{\tabcolsep}{5pt}
\resizebox{\linewidth}{!}{
\begin{tabular}{@{\extracolsep{4pt}}cc c  cc@{} }
\toprule

\textbf{\multirow{2}{*}{Variant}}  
& \multirow{2}{*}{\shortstack[c]{Textual \\ Alignment $S_{\text{CLIP}}^\mathrm{T}$ ($\uparrow $) }}
&  \multicolumn{2}{c}{Concept Alignment}  \\
\cmidrule{3-4} 
 && $S_{\text{CLIP}}^\mathrm{I}$ ($\uparrow $) & $S_{\text{DINO}}$ ($\uparrow $) \\

\midrule[0.5pt]
\textbf{w/o base flow} & $\textbf{0.3718}$ & $\text{0.5861}$ & $\text{0.4337}$ \\
\textbf{w/o prompt variety} & $\text{0.3539}$ & $\text{0.7155}$ & $\text{0.5648}$ \\

\midrule

\textbf{\latexblend (Ours)} & $\text{0.3684}$ & $\textbf{0.8052}$ & $\textbf{0.6564}$ \\

\bottomrule
\vspace{-10pt}
\end{tabular}
}
\vspace{-8pt}
\caption{\textbf{Ablation study.} 
Both the base flow and prompt variety play crucial roles in obtaining an effective and robust concept representation that encompasses sufficient concept information.
}

\label{tbl:ablation}
\vspace{-10pt}
\end{table}

\myparagraph{Blending Guidance.}
We conduct ablation experiments to verify the effect of blending guidance on multi-concept inference, with visual examples provided in Fig.~\ref{fig:ablation2}.
In the \emph{``w/o blending guidance"} scenario, we merge multiple concepts by performing direct feature replacement in the latent textual space (Eq.~\ref{multi_merge}).
We observe that this straightforward merging strategy performs well in most cases when the number of concepts is small (typically $\leq 3$), with blending guidance offering only subtle improvement. 
As the number of concepts increases, blending guidance greatly enhances multi-concept inference, mitigating issues such as concept fidelity degradation and identity mixing.

\section{Conclusion}
\label{sec:conclusion}

In conclusion, \latexblend is a novel framework that effectively and efficiently scales multi-concept customized text-to-image generation by representing and blending concepts in a latent space of textual features. 
Qualitative and quantitative results demonstrate that \latexblend significantly outperforms previous methods in both generation quality and computational efficiency. 
\latexblend greatly alleviates denoising deviation in customized generation and preserves the coherent structure arranged by the pre-trained T2I model.
However, current T2I diffusion models may struggle with lengthy and complex prompts describing intricate scenes with multiple objects~\cite{bakr2023hrs, gani2023llm}.
This inherent limitation of pre-trained models hinders the further scaling of concept numbers in multi-concept inference.
We hope our findings can shed light on future advancements in customized visual content creation.

\paragraph{Acknowledgment.}
This work was supported by the National Science Fund of China under Grant Nos. U24A20330, 62361166670, 62276132, and 61876085.

{
    \small
    \bibliographystyle{ieeenat_fullname}
    \bibliography{main}
}


\appendix
\renewcommand{\thefootnote}{\arabic{footnote}}

\clearpage
\noindent{\Large\bf Appendix}
\vspace{5pt}

The supplementary material is organized as follows:
In \refsec{exp_res}, we provide additional experimental results, including further ablation studies, comparisons of single-concept customized generation, and more comparisons with baseline methods. 
In \refsec{imp_deta}, we provide the implementation details of our method and the baselines. 
In \refsec{societal}, we discuss the societal impacts of our work.

\section{More Experimental Results}\lblsec{exp_res}

\subsection{More Ablation Studies}

\begin{figure}[!t]
 \centering
 \includegraphics[width=\linewidth]{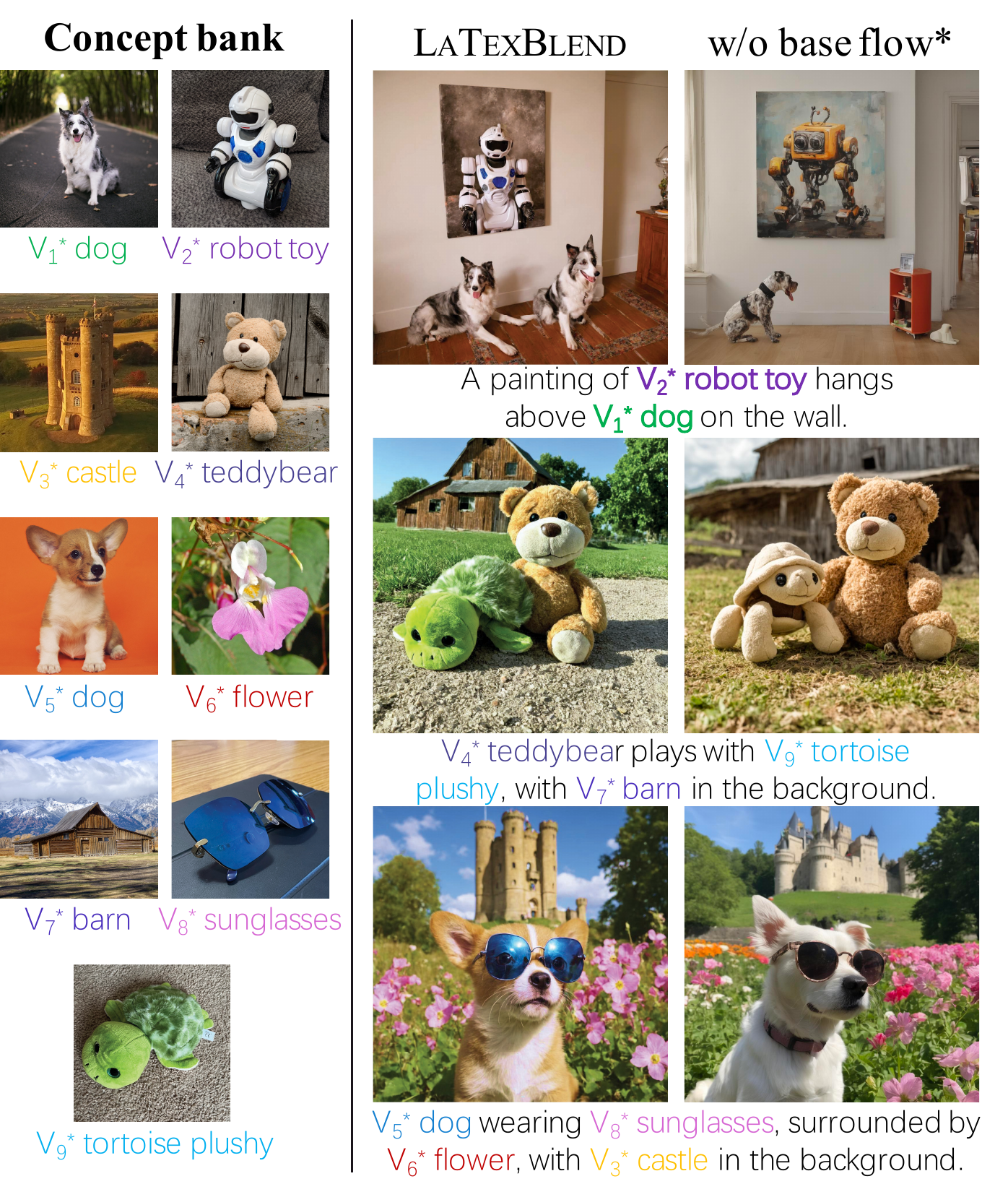}
 \caption{\textbf{Additional ablation study on the base encoding flow.} 
 The base encoding flow remains necessary when the fine-tuning prompt contains only concept-related tokens. Without it, the obtained $\mathbf{h}_c$ lacks sufficient concept-related information, which may potentially disperse into the padding tokens.
 }
 \label{fig:simple_temp}
\end{figure}

\paragraph{Additional Ablation Study on Base Encoding Flow.}
We explore whether concept-related information is dispersed across latent textual features when the fine-tuning prompt contains only concept-related tokens. 
Specifically, the prompt takes the form of {\menlo "V* <noun>"}, where {\menlo "V*"} is an identifier token and {\menlo "<noun>"} is a coarse class descriptor of the subject (\emph{e.g.} , cat, dog).
To validate the necessity of the base encoding flow in this specific scenario, we conduct additional ablation studies. In the \emph{``w/o base flow*"} scenario, the base flow is removed, and fine-tuning is performed using only the single learnable concept flow $\mathscr{F}_c$. 
Besides, the fine-tuning prompt adopts the form {\menlo "V* <noun>"} and is padded to a fixed sequence length $M$. 
We show sample generations in Fig.~\ref{fig:simple_temp}.
As demonstrated, the generated images exhibit significant degradation in concept fidelity in the \emph{``w/o base flow*"} scenario. This result suggests that the obtained $\mathbf{h}_c$ lacks sufficient concept-related information, potentially dispersing into the padding tokens. 
It further confirms the necessity of the base encoding flow for obtaining an effective concept representation.

\begin{figure*}[!t]
 \centering
 \includegraphics[width=\linewidth]{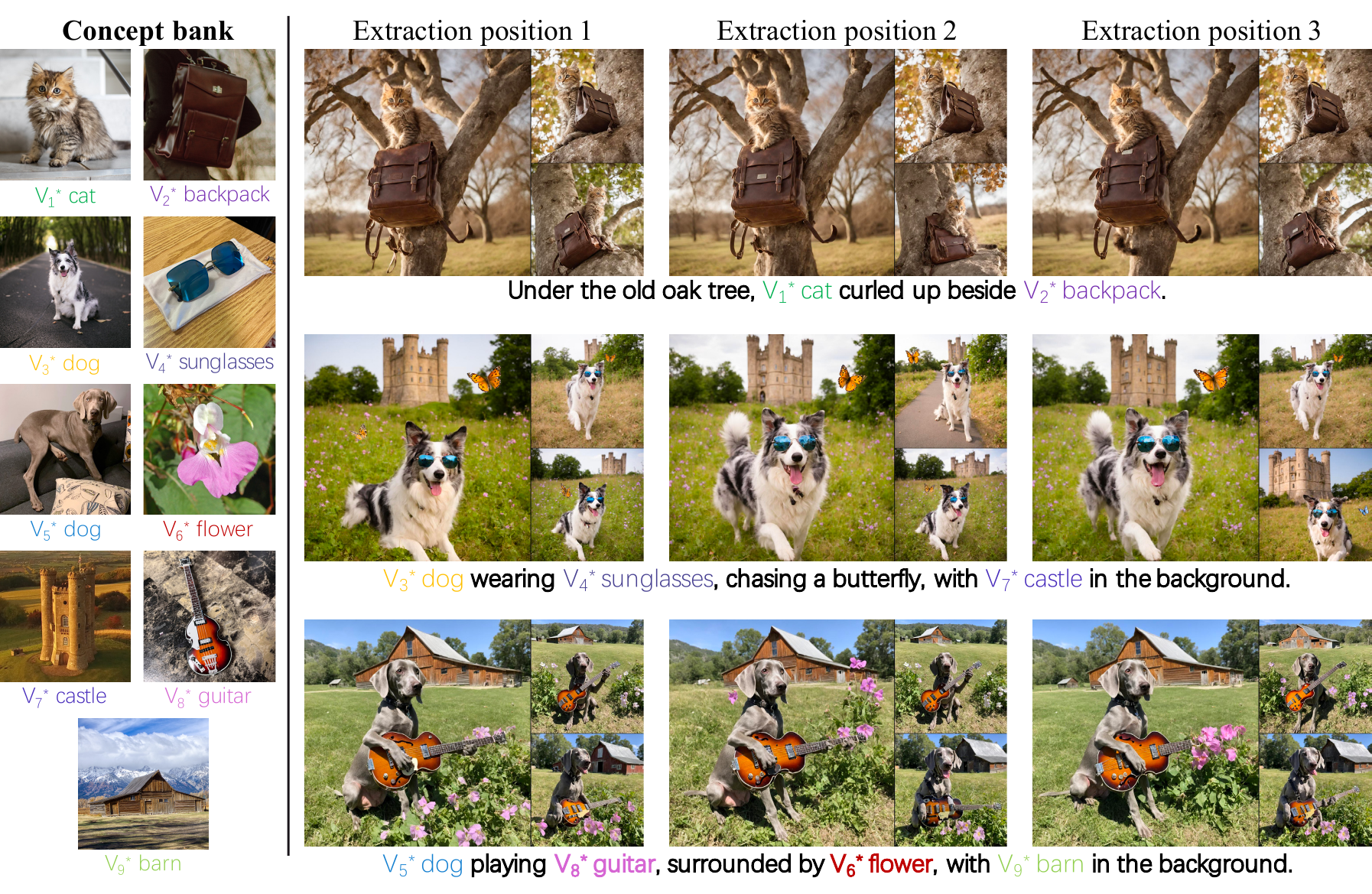}
 \caption{\textbf{Position invariance.} 
 Different columns use $\mathbf{h}_c$ extracted from different prompt templates. 
 For each generation case, we present three images generated with different noise initializations.
 Although the representations $\mathbf{h}_c$ used in different columns are extracted from different templates, they function correctly regardless of their extraction positions and produce similar results with the same noise initialization and prompt.
 }
 \label{fig:pos_inv}
\end{figure*}

\paragraph{Position Invariance}\label{pos_inv}
At inference, the blending position of $\mathbf{h}_c$ can vary throughout the prompt. 
Therefore, we employ a prompt variation strategy for single-concept customization in \latexblend, dynamically varying the prompt template to construct textual prompts for the two textual encoding flows.
By varying the extraction and insertion positions of $\mathbf{h}_c$, we aim to eliminate its positional dependency.
We present sample generations in Fig.~\ref{fig:pos_inv} to illustrate the position invariance of $\mathbf{h}_c$. 
Different columns use $\mathbf{h}_c$ obtained from different prompt templates. 
Specifically, the $\mathbf{h}_c$ used in columns 1, 2, and 3 are extracted from the templates $\texttt{"\{\}."}$, $\texttt{"Photo of \{\}."}$, and $\texttt{"A fancy photo of \{\}."}$, respectively.
For each generation case, we present three images generated with different noise initializations.
As observed, although the representations $\mathbf{h}_c$ used in different columns are extracted from different templates, they function correctly regardless of their extraction positions and produce similar results with the same noise initialization and prompt.
Furthermore, the invariance of $\mathbf{h}_c$ with respect to its extraction position remains robust when its blending position varies.
In our experiment, all concept representations $\mathbf{h}_c$ are extracted from the prompt template $\texttt{"Photo of \{\}."}$.

\subsection{Single-concept Customized Generation}\lblsec{single_concept}

\begin{figure*}[!t]
 \centering
 \includegraphics[width=\linewidth]{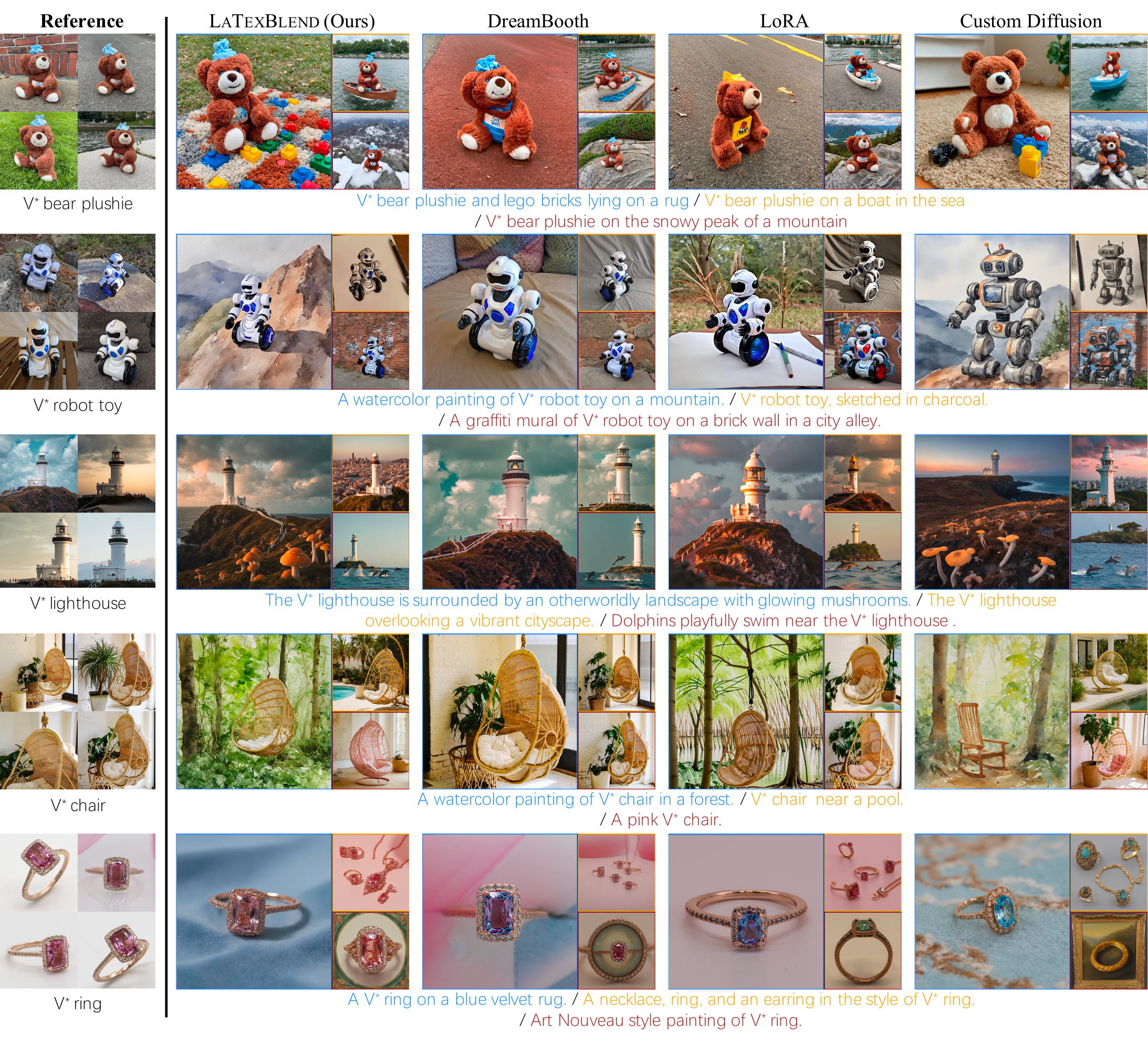}
 \caption{\textbf{Visual comparison of single-concept customized generation.} 
 We set the rank of LoRA to 4 by default. DreamBooth often overfits, resulting in a loss of editability, while Custom Diffusion struggles to maintain concept fidelity.
 \latexblend generates the customized subject with high concept fidelity while faithfully adhering to the query prompt.
 }
 \label{fig:single_concept}
\end{figure*}

The proposed \latexblend can also be applied to single-concept customized generation, where a single concept representation is blended with the output of the base encoding flow in the latent textual space:
\begin{equation}
    \mathcal{F}(\mathbf{h}_\mathcal{C}) = \texttt{Blend}(\mathbf{h}_b; \mathbf{h}_{c_1})\,,
\end{equation}
where $\mathbf{h}_b$ denotes the output of the base encoding flow, while $\mathbf{h}_{c_1}$ is the representation of the target concept.
We compare \latexblend with several representative single-concept customized generation methods, including DreamBooth~\cite{ref2}, Custom Diffusion~\cite{kumari2023multi}, and LoRA~\cite{hu2021lora}. Sample generations are presented in Fig.~\ref{fig:single_concept}.
As observed, DreamBooth fine-tunes the entire U-Net, which often leads to overfitting and a loss of editability, making it difficult to accurately render the target subject within the query context. 
Custom Diffusion falls short in concept fidelity.
In contrast, \latexblend generates the customized subject with high concept fidelity while faithfully adhering to the query prompt. This further validates the effectiveness of \latexblend in representing concepts and mitigating denoising deviation in customized generation.

\subsection{Comparison with More Baselines}
\begin{figure*}[!t]
 \centering
 \includegraphics[width=\linewidth]{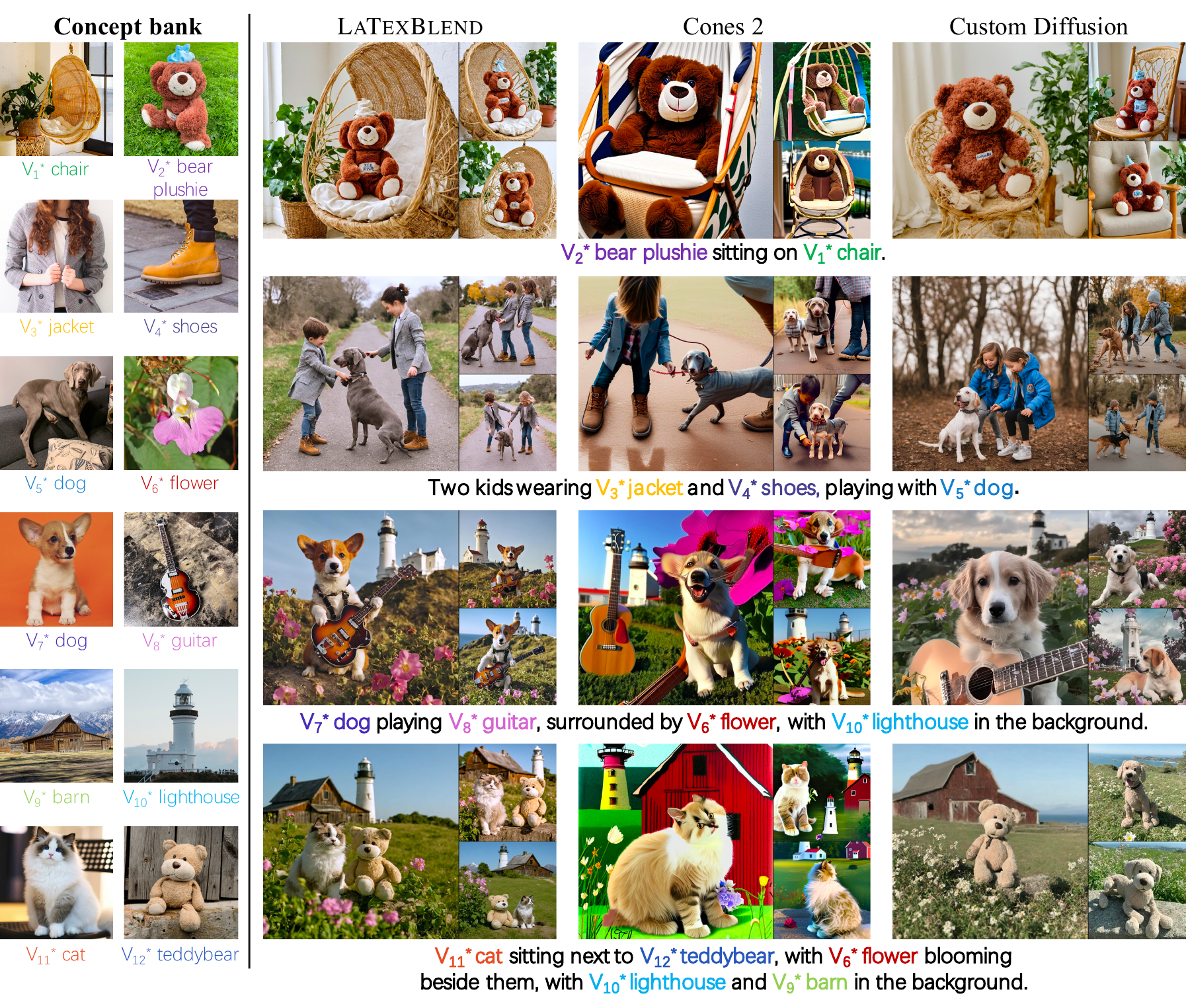}
 \caption{\textbf{Comparison with more multi-concept customized generation Methods.} 
 We perform qualitative comparisons of \latexblend with two earlier multi-concept customized generation methods, including Cones 2~\cite{liu2023cones2} and Custom Diffusion~\cite{kumari2023multi}.
 \latexblend demonstrates advantages over the baselines in both subject fidelity and image structure coherence.
 }
 \label{fig:more_baseline}
\end{figure*}

\lblsec{poor_performance}
Due to space limitations, we omit visual comparison with some earlier multi-concept customized generation methods in the paper.
Therefore, we supplement the qualitative comparisons of \latexblend with these methods, including Cones 2~\cite{liu2023cones2} and Custom Diffusion~\cite{kumari2023multi}.
For each method, we randomly generate 10 images per case and select the best 3 for visual comparison. 
We present sample generations in Fig.~\ref{fig:more_baseline}.
As observed, Cones 2 relies heavily on additional layout conditioning. In the absence of predefined layout conditions, Cones 2 suffers from degradation in both subject fidelity and image structure. 
Images generated with explicit layout conditions lack diversity in their layouts and may struggle to capture complex semantics, such as inter-subject interactions and actions.
Custom Diffusion falls short in maintaining subject fidelity and ensuring coherence in image structure.

Guidance~\cite{ho2022classifier, dhariwal2021diffusion} is a commonly used technique in conditional image generation, which adjusts the sampling process toward specific targets by modifying the update rule of noisy latents.
We propose blending guidance in \latexblend, which leverages mutual information from different concepts within a single denoising branch to rectify attention.
$\text{MC}^2$~\cite{jiang2024mc} is a guidance-based multi-concept customized generation method.
Unlike \latexblend, $\text{MC}^2$ derives guidance from attention relations across multiple denoising branches.
Therefore, its inference-time computation is high and scales proportionally with the number of concepts.

\subsection{Comparison with Learning-based Method}
\begin{figure*}[!t]
 \centering
 \includegraphics[width=0.9\linewidth]{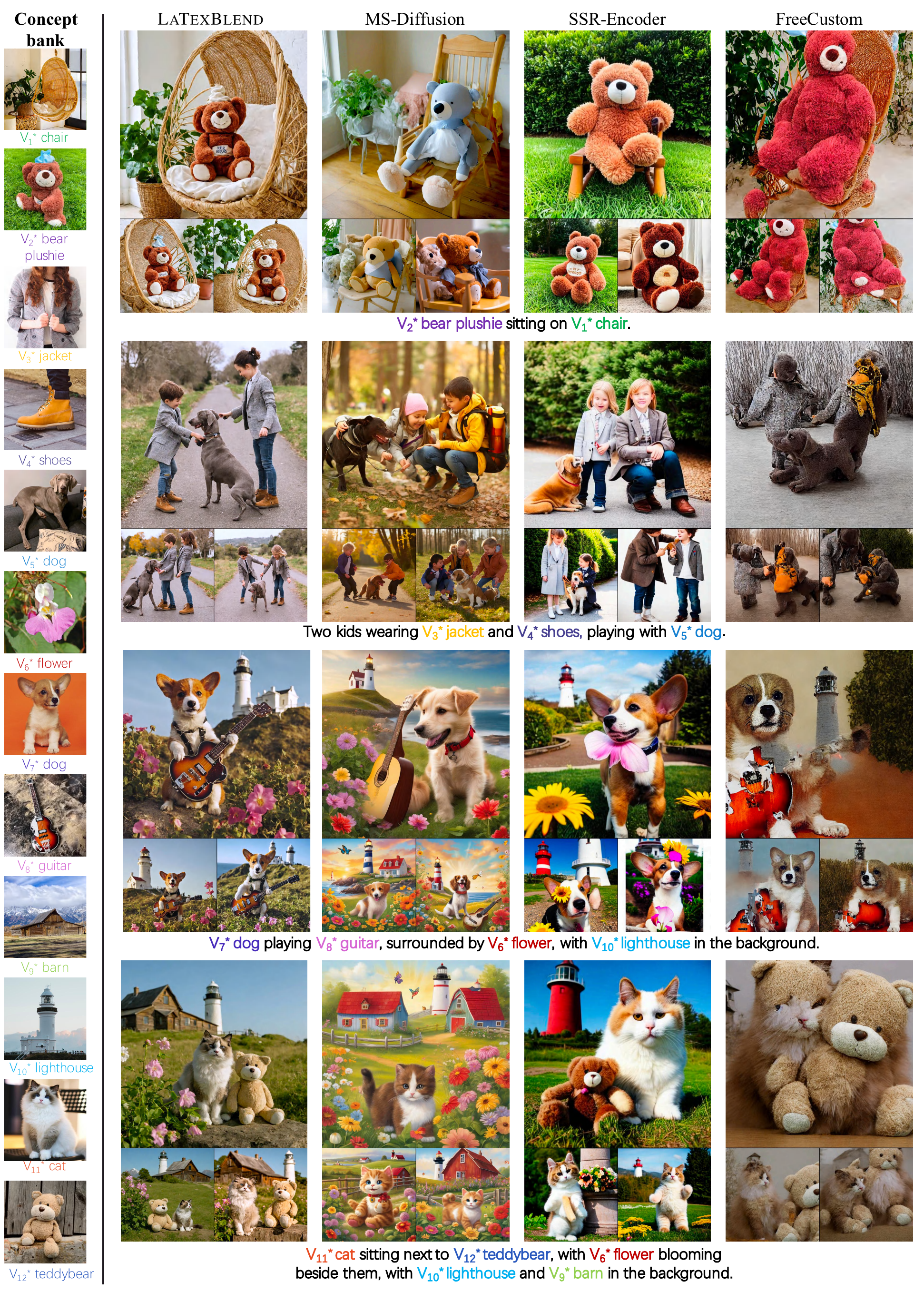}
 \vspace{-10pt}
 \caption{\textbf{Comparison with learning-based multi-concept customized generation methods.} 
 We compare our approach with several existing multi-concept learning-based methods, including FreeCustom~\cite{ding2024freecustom}, SSR-Encoder~\cite{zhang2024ssr}, and MS Diffusion~\cite{wang2024ms}.
 The primary issue with learning-based methods is their struggle to generate customized subjects with high subject fidelity, particularly for complex subjects.
 }
 \label{fig:training_free}
 \vspace{-40pt}
\end{figure*}

\lblsec{learning_based}
Apart from optimization-based methods, another line of research on customized generation is learning-based method~\cite{wei2023elite, ding2024freecustom, zhang2024ssr, wang2024ms}, which aims to train unified models capable of personalizing diverse subject inputs.
We compare our approach with several existing learning-based multi-concept customized generation methods, including FreeCustom~\cite{ding2024freecustom}, SSR-Encoder~\cite{zhang2024ssr}, and MS Diffusion~\cite{wang2024ms}.
We utilize the official implementation for FreeCustom\footnote{\url{https://github.com/aim-uofa/FreeCustom}}, 
SSR-Encoder\footnote{\url{https://github.com/Xiaojiu-z/SSR_Encoder}}, 
and MS Diffusion\footnote{\url{https://github.com/MS-Diffusion/MS-Diffusion}}.
For each competing method, we randomly generate 10 images per case and select the best 3 for visual comparison. 
The results of the qualitative comparison are shown in Fig.~\ref{fig:training_free}. 
As we can see, these learning-based methods face challenges in faithfully generating target subjects and preserving their key identifying features, resulting in low subject fidelity - especially for complex customized subjects. 

\subsection{Supplementary Experimental Results}\lblsec{sup_res}

\begin{figure*}[!t]
 \centering
 \includegraphics[width=\linewidth]{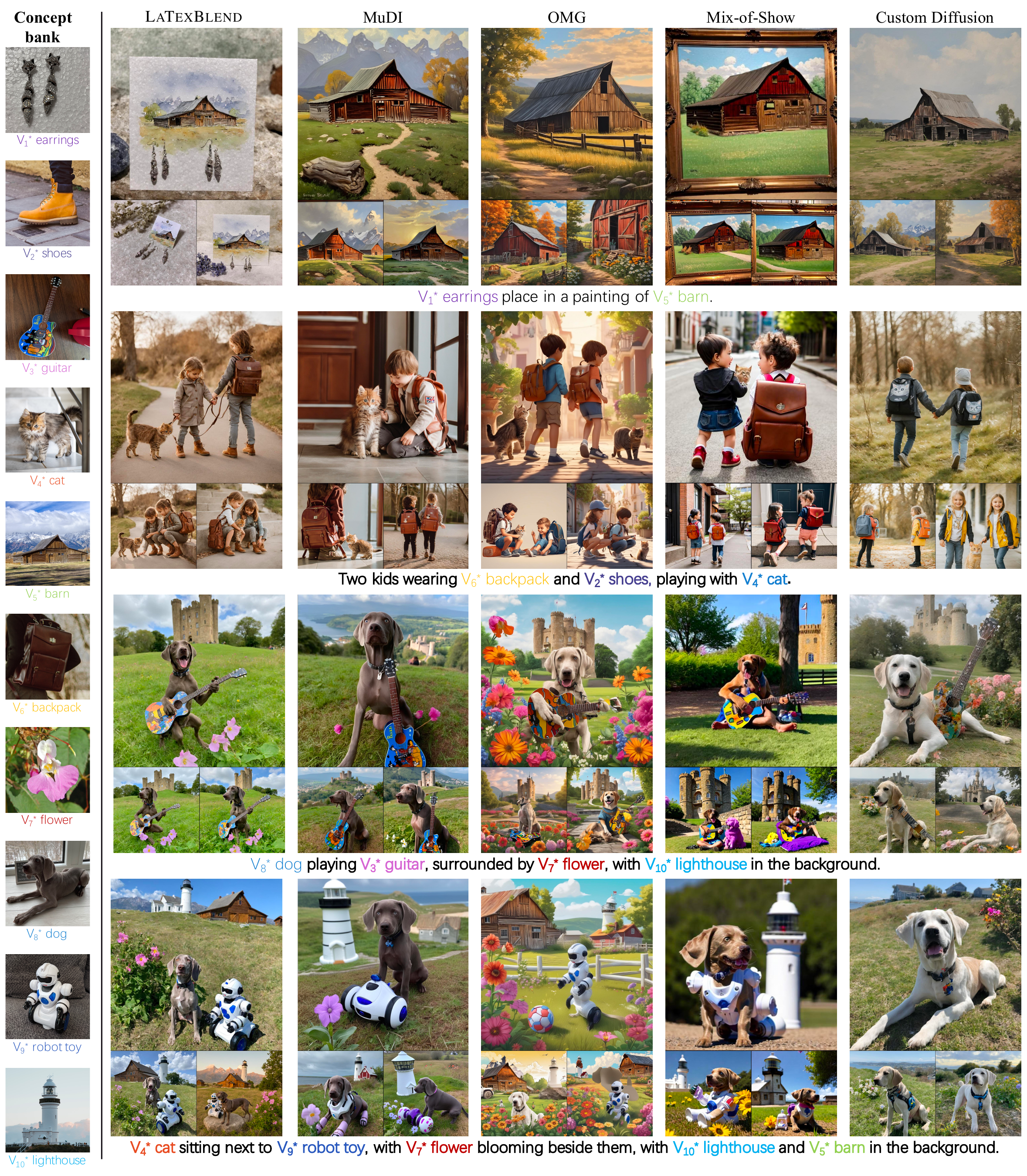}
 \vspace{-20pt}
 \caption{\textbf{More visual comparisons with baselines.} 
  We provide more qualitative comparisons of multi-concept generation between \latexblend and baseline methods, including Custom Diffusion~\cite{kumari2023multi}, Mix-of-Show~\cite{gu2024mix}, OMG~\cite{kong2024omg}, and MuDI~\cite{jang2024identity}.
 }
 \vspace{-10pt}
 \label{fig:more_comp}
\end{figure*}

\paragraph{More Results of Visual Comparison.}
We provide more qualitative comparisons of multi-concept generation between \latexblend and baseline methods, including Custom Diffusion~\cite{kumari2023multi}, Mix-of-Show~\cite{gu2024mix}, OMG~\cite{kong2024omg}, and MuDI~\cite{jang2024identity}, as shown in Fig.~\ref{fig:more_comp}. 
For each method, we randomly generate 10 images per case and select the best 3 for visual comparison.
MuDI, Mix-of-Show, and Custom Diffusion exhibit issue of image structure degradation, often producing single-object-centric subjects or omitting target subjects. 
OMG's performance heavily relies on the accuracy of the segmentation model~\cite{kirillov2023segment}, occasionally failing to integrate customized subjects, which results in gray-shaded areas. 
Besides, OMG struggles to maintain overall consistency and realism in image style. 

\paragraph{Detailed Data of Quantitative Comparison.}
The detailed data of Fig. 7 in the paper is provided in~\reftbl{detailed_data}.
As shown, the proposed \latexblend significantly outperforms all baseline methods in concept alignment, particularly in terms of the DINO score.
Compared with CLIP, DINO can better capture the unique features of each subject, thereby better reflecting fine subject similarity rather than coarse class similarity~\cite{ref2}.
The superiority in DINO score highlights \latexblend's ability to effectively preserve the key identifying features of target subjects.

\begin{table*}[!t]
\centering
\setlength{\tabcolsep}{5pt}
\resizebox{\linewidth}{!}{
\begin{tabular}{cl cccccc | c }
\toprule

& \multirow{2}{*}{\textbf{Methods}}
& \multirow{2}{*}{\shortstack[c]{One concept}}
& \multirow{2}{*}{Two concepts}
& \multirow{2}{*}{\shortstack[c]{Three concepts}}
& \multirow{2}{*}{\shortstack[c]{Four concepts}}
& \multirow{2}{*}{\shortstack[c]{Five concepts}}
& \multirow{2}{*}{\textbf{Average} }
& \multirow{2}{*}{\shortstack[c]{Five concepts \\ w layout}}
\\ \\

\midrule

\multirow{5}{*}{\shortstack[c]{Concept-alignment \\ CLIP score}} &

\textbf{Cones 2} & 0.7124 & 0.7079 & 0.737 & 0.7219 & 0.6988 & 0.7155 & \underline{0.7614} \\
& \textbf{Mix-of-Show} & 0.7222 & 0.6785 & 0.6846 & 0.6766 & 0.7343 & 0.6992 & 0.7067 \\
& \textbf{OMG} & 0.7786 & 0.7405 & 0.6834 & 0.7067 & 0.6974 & 0.7213 & -\\
& \textbf{MuDI} & 0.7415 & 0.7131 & 0.7232 & 0.7379 & 0.7641 & \underline{0.7359} & 0.7211  \\
& \textbf{\latexblend (Ours)} & 0.7766 & 0.7322 & 0.7688 & 0.7553  & 0.795 & \textbf{0.7656} & \textbf{0.7829} \\

\midrule
\multirow{5}{*}{\shortstack[c]{Concept-alignment \\ DINO score}} &
\textbf{Cones 2} & 0.3494 & 0.3903 & 0.4151 & 0.4286 & 0.363 & 0.3893 & \underline{0.4278} \\
& \textbf{Mix-of-Show} & 0.4838 & 0.4817 & 0.4055 & 0.395 & 0.3915 & 0.4315 & 0.3886 \\
& \textbf{OMG} & 0.4914 & 0.5213 & 0.4994 & 0.4405 & 0.4874 & 0.488 & -  \\
& \textbf{MuDI} & 0.5345 & 0.4902 & 0.4995 & 0.5002 & 0.488 & \underline{0.5025} & 0.3826  \\
& \textbf{\latexblend (Ours)} & 0.5846 & 0.5892 & 0.5729 & 0.4922 & 0.5196 & \textbf{0.5517} & \textbf{0.5214} \\

\midrule 
\multirow{5}{*}{\shortstack[c]{Text-alignment}} & 

\textbf{Cones 2} & 0.3857 & 0.3201 & 0.3205 & 0.2959 & 0.3899 & 0.3424 & 0.3193 \\
& \textbf{Mix-of-Show} & 0.334 & 0.3101 & 0.2817 & 0.3224 & 0.3893 & 0.3275 & \textbf{0.353}\\
& \textbf{OMG} & 0.3616 & 0.3433 & 0.3215 & 0.4299 & 0.3816 & \textbf{0.3675} & - \\
& \textbf{MuDI} & 0.35  & 0.2817 & 0.3035 & 0.3805 & 0.3687 & 0.3368 & 0.3109 \\
& \textbf{\latexblend (Ours)} & 0.3745 & 0.327 & 0.3025 & 0.4242 & 0.4044 & \underline{0.3665} & \underline{0.348}\\

\bottomrule
\end{tabular}}
\vspace{-8pt}
\caption{\textbf{Detailed data of quantitative evaluation on multi-concept generation}.
 We highlight the best result in bold and underline the second best for different settings.
\latexblend outperforms all baseline methods in concept alignment and demonstrates competitive performance in prompt fidelity. The clear advantages of \latexblend in the DINO score showcase its ability to effectively preserve the key identifying features of target subjects.
}

\label{tbl:detailed_data}
\vspace{-8pt}
\end{table*}

\section{Implementation Details}\lblsec{imp_deta}

\subsection{Subjects and Prompts}

\begin{figure}[!t]
 \centering
 \includegraphics[width=\linewidth]{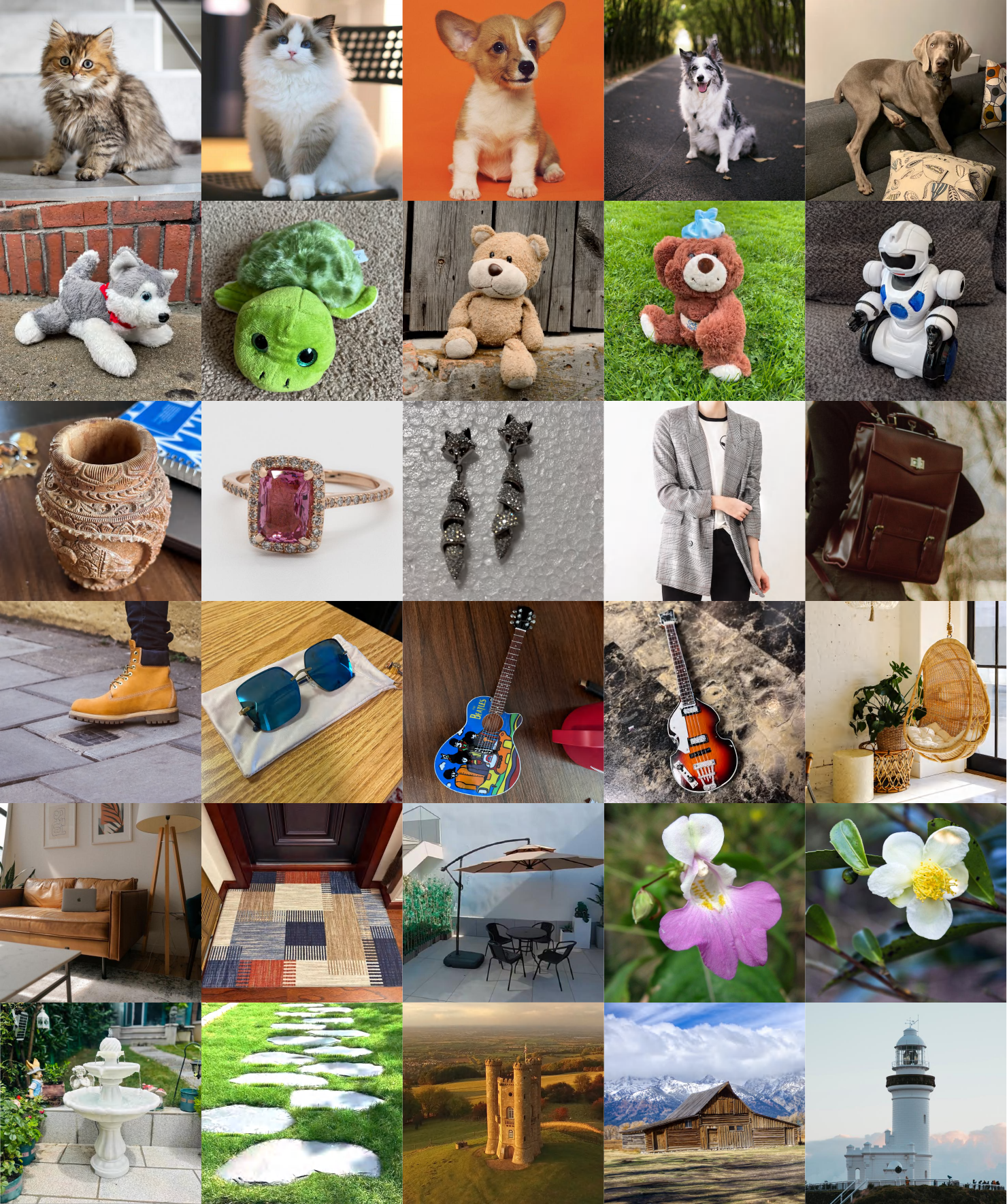}
 \caption{\textbf{One reference image of each subject.} 
 }
 \label{fig:concept_bank}
\end{figure}

\begin{figure}[!t]
 \centering
 \includegraphics[width=\linewidth]{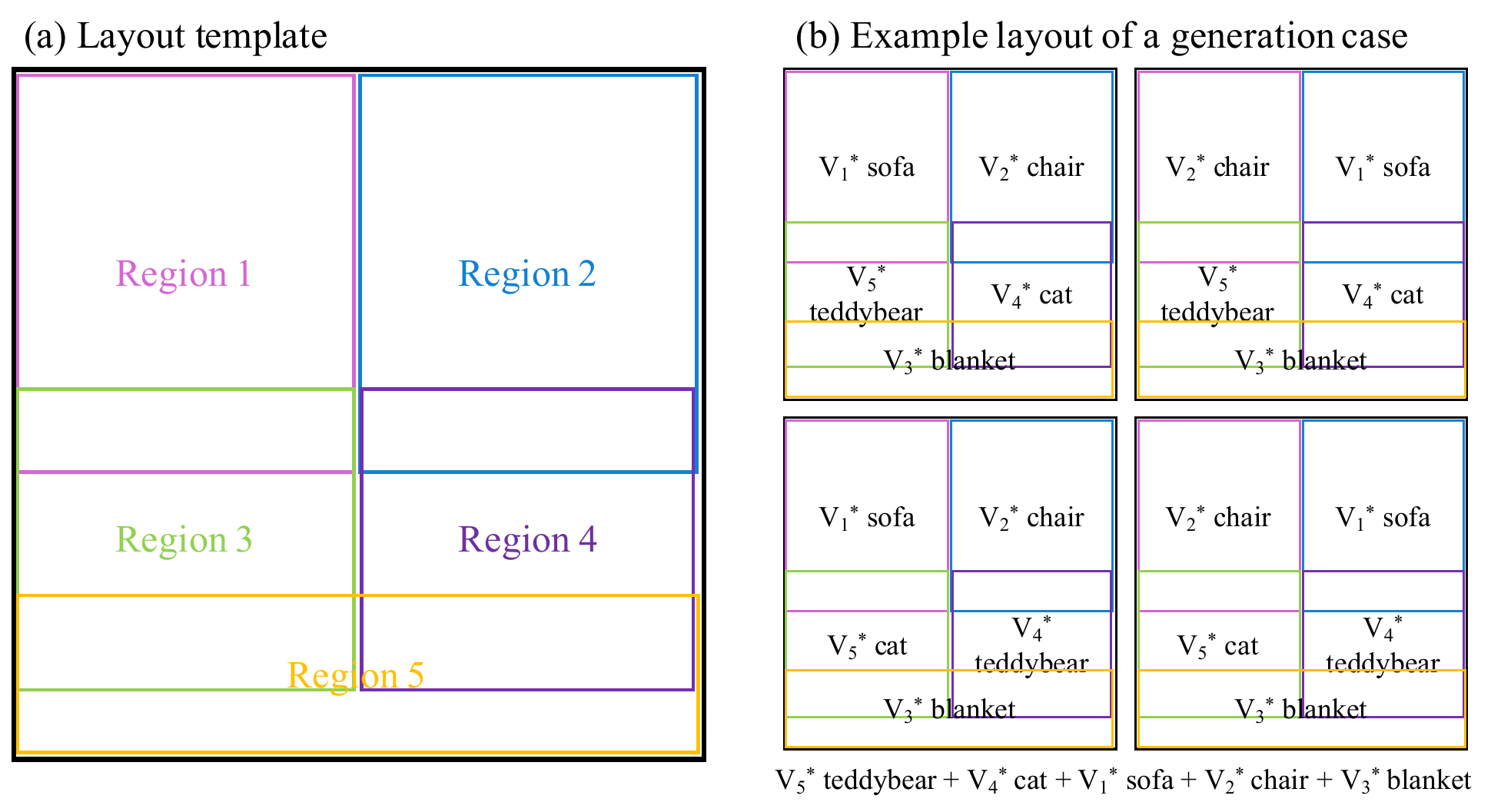}
 \caption{\textbf{Layout template for layout-conditioned multi-concept generation.} 
 \textbf{(a).} The layout template.
 \textbf{(b).} For each generation case, we sequentially swap the specified subjects in regions 1 and 2, and those in regions 3 and 4, resulting in a total of 4 different layout instances.
 }
 \label{fig:layout}
\end{figure}

We conduct experiments on 30 different subjects, most of which are sourced from previous studies~\cite{ref6, ref2, kumari2023multi}. 
Additionally, we collect images of some new subjects missing in previous studies.
These subjects cover various categories, such as buildings, pets, and objects, each represented by several reference images.
We show one reference image of each subject in Fig.~\ref{fig:concept_bank}.
The subject combinations and corresponding query prompts are initially generated by ChatGPT and subsequently reviewed and curated manually.
Specifically, we create a $\texttt{concept list}$ containing the 30 subjects mentioned above, then generate $n$-concept subject combinations and their query prompts using ChatGPT with the following several steps (where $n$ is replaced by the specific number): 1. Select $n$ subjects from the $\texttt{concept list}$.
2. Use the selected $n$ nouns to construct sentences based on the following requirements: 1) Apart from the $n$ selected nouns, avoid using other nouns from the concept list in the sentence. 2) Ensure the sentences are logical. 3) Maintain diversity in the sentence structures.
Some prompts are also inspired by other previous works~\cite{kumari2023multi, ref2}.

\begin{figure*}[!t]
 \centering
 \includegraphics[width=0.7\linewidth]{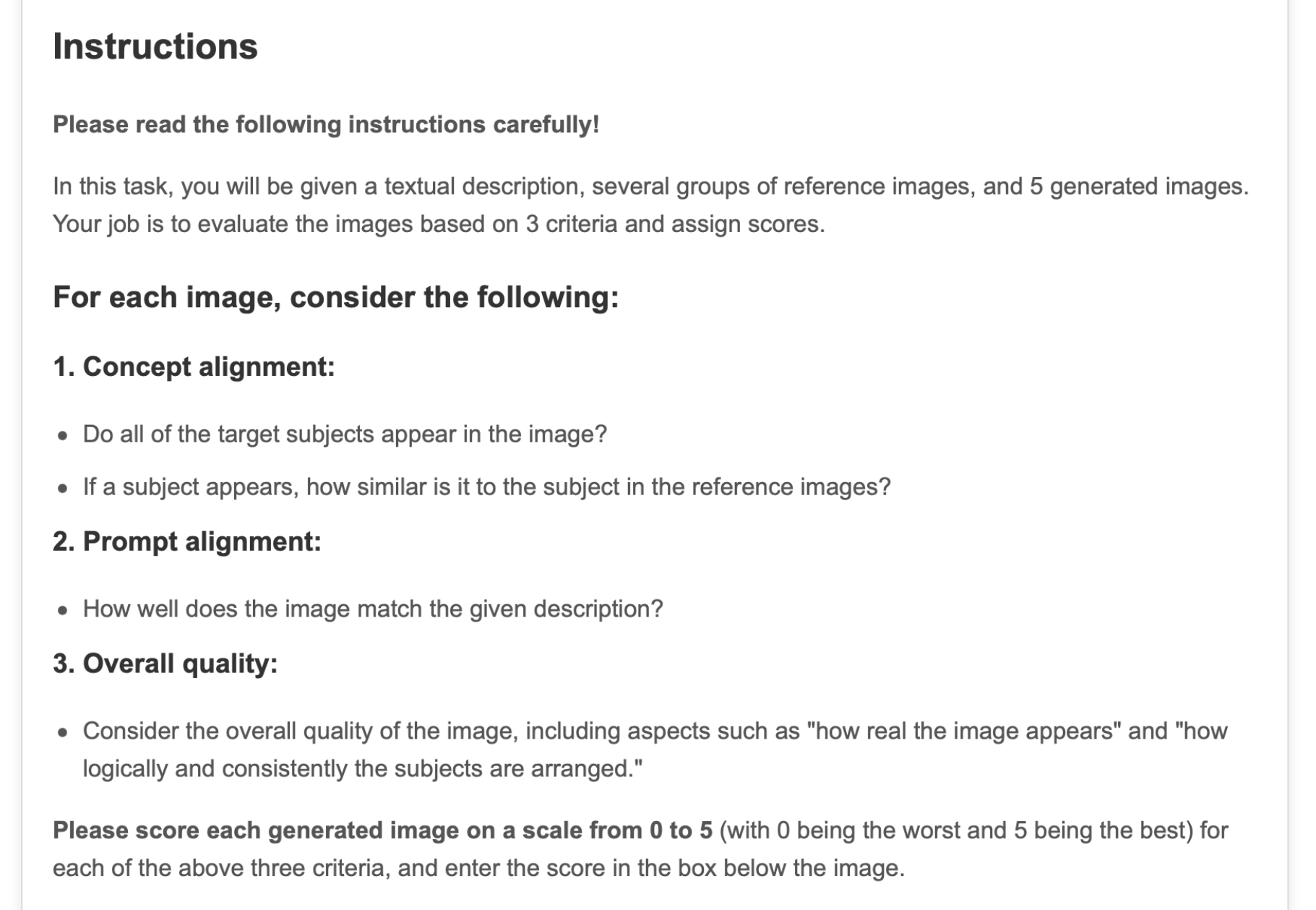}
 \caption{\textbf{The instructions that were given to the participants.} 
 }
 \label{fig:instruct}
\end{figure*}

\begin{figure*}[!t]
 \centering
 \includegraphics[width=0.9\linewidth]{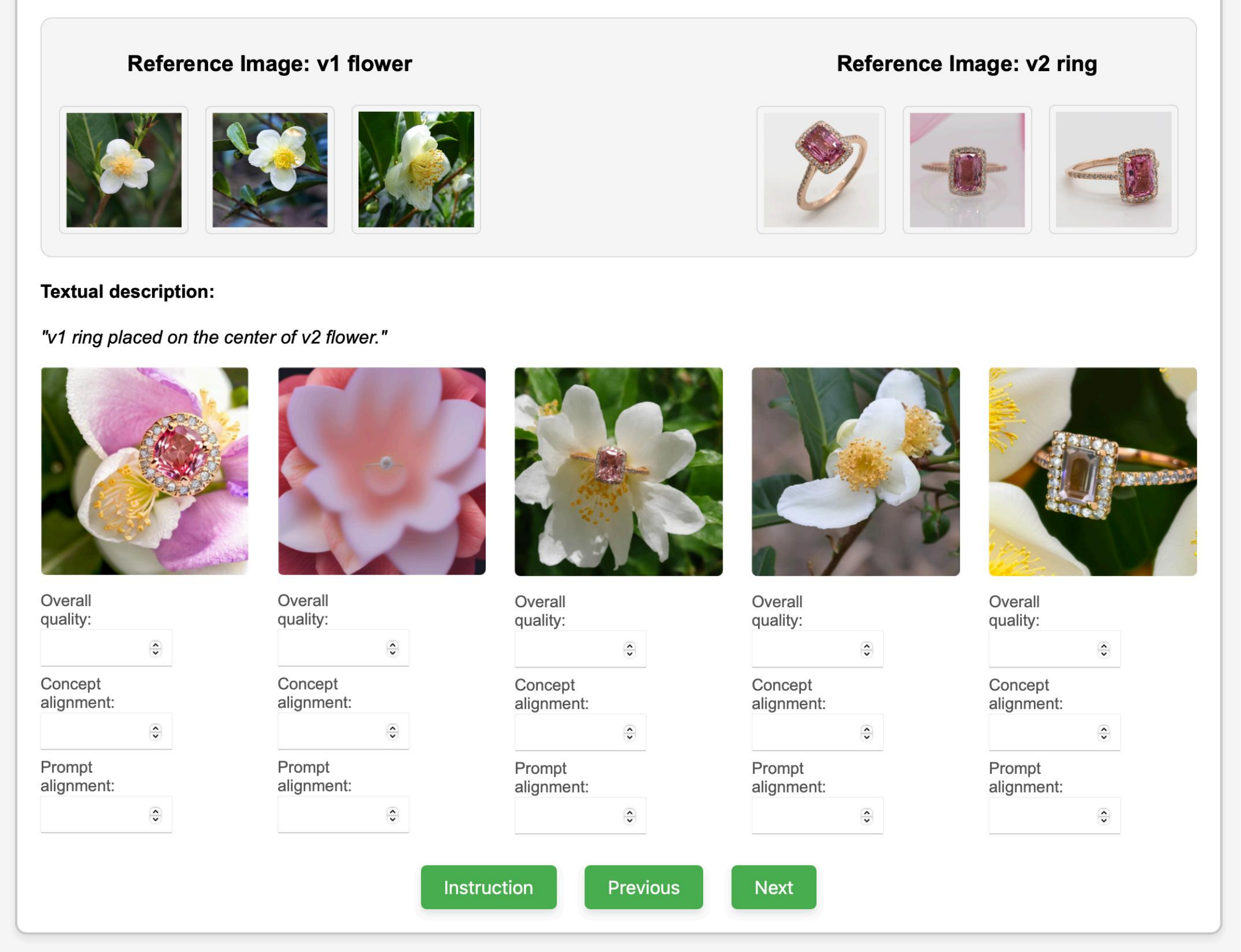}
 \caption{\textbf{A screenshot of an evaluation case from the user study.} 
 }
 \label{fig:user_exp}
\end{figure*}

\subsection{Prompt Template Pool}\label{temp_pool}
For the customization of each single concept in \latexblend, we create a template pool containing 7 different prompt templates. The complete list of templates is:
\begin{verbatim}
1. "{}."
2. "A {}."
3. "Photo of {}."
4. "A photo of {}."
5. "A photo of a {}."
6. "a fancy photo of a {}."
7. "A fancy, detailed photo of {}."
\end{verbatim}
During fine-tuning, we randomly draw different templates from the prompt template pool to construct prompts for the two textual encoding flows. At inference, the compact representation $\mathbf{h}_c$ of each concept is obtained from the template $\texttt{"Photo of \{\}."}$.

\subsection{Layout Conditioning}\lblsec{layout}
We conduct experiments of multi-concept customized generation with additional layout conditioning in Section 4.2 of the paper.
The layout template we use is shown in Fig.~\ref{fig:layout} (a). 
There are 5 specified generation regions in the layout template for 5-concept generation.
For each generation case, we alternately swap the specified subjects in regions 1 and 2, and those in regions 3 and 4, resulting in a total of 4 different layout instances.
We also provide an example of all 4 layout instances for a generation case in Fig.~\ref{fig:layout} (b).
For each generation case, we randomly generate 5 images per layout instance using different methods and select the best 3 from the resulting 20 images for visual comparison.

\subsection{Additional Details on User Study}\lblsec{user_study}

We conduct a user study with 25 participants, using 20 sets of generation cases with the number of concept ranging from 2 to 5.
For each generation case, we randomly generate 5 images per query prompt using each competing method to create an image candidate pool. 
Before evaluation, participants are thoroughly briefed on the scoring rules and provided with scoring examples.
We provide a screenshot of the instructions given to participants in Fig.~\ref{fig:instruct}.
The images from Cones 2 in the user study are generated without additional layout conditioning for a fair comparison. 
In each evaluation case, participants are given a textual prompt, reference images of the customized subject, and corresponding generations from different methods.
Generated images are randomly selected from the image candidate pool and presented side-by-side in a random order to participants. Participants are given unlimited time to score each generation on a scale from 0 to 5 (with 0 being the worst and 5 the best) based on three criteria:
1) whether the image contains all target subjects and aligns with their visual appearance in the reference images,
2) whether the image content adheres to the scenes described by the textual prompt, and 3) the overall quality in terms of authenticity and coherence.
A screenshot of an evaluation case is provided in Fig.~\ref{fig:user_exp}.

\subsection{Implementation Details}\lblsec{imp_base}
In our experiment, we use Stable Diffusion XL (SDXL)~\cite{podellsdxl} as the pre-trained text-to-image diffusion model.
Images are generated using 100 DDIM sampling steps with a classifier-free guidance scale of 6 for all compared methods.
All model fine-tuning is conducted on NVIDIA GeForce RTX 4090 GPUs, and inference is conducted on a 40GB NVIDIA A100 GPU.

\paragraph{Custom Diffusion.}
We employ the official implementation\footnote{\url{https://github.com/adobe-research/custom-diffusion}} for Custom Diffusion~\cite{kumari2023multi}. 
The model is fine-tuned using the default hyperparameters and settings provided in the code. 
Custom Diffusion requires joint training for multi-concept generation. 
The recommended fine-tuning steps are 500 for a single concept and 1000 for two concepts; accordingly, we increase the training steps by 500 for each additional concept.

\paragraph{Mix-of-Show.}
We use the official implementation\footnote{\url{https://github.com/TencentARC/Mix-of-Show}} of Mix-of-Show~\cite{gu2024mix}. 
Following the authors' guidelines and examples, we make extra annotations for the reference images, including subject masks and detailed image captions.
For single-concept fine-tuning, each concept is associated with two identifier tokens, represented in the form {\menlo "[V1] [V2] <noun1>"}.
We fine-tune the single-concept model and fuse multiple models using the default parameters provided in the official implementation. The fusion operation is performed for each distinct subject combination.
For multi-concept generation with layout conditions, we manually create sketch-based conditions in the form of precise object contours, following the format of the official examples.

\begin{figure}[!t]
 \centering
 \includegraphics[width=\linewidth]{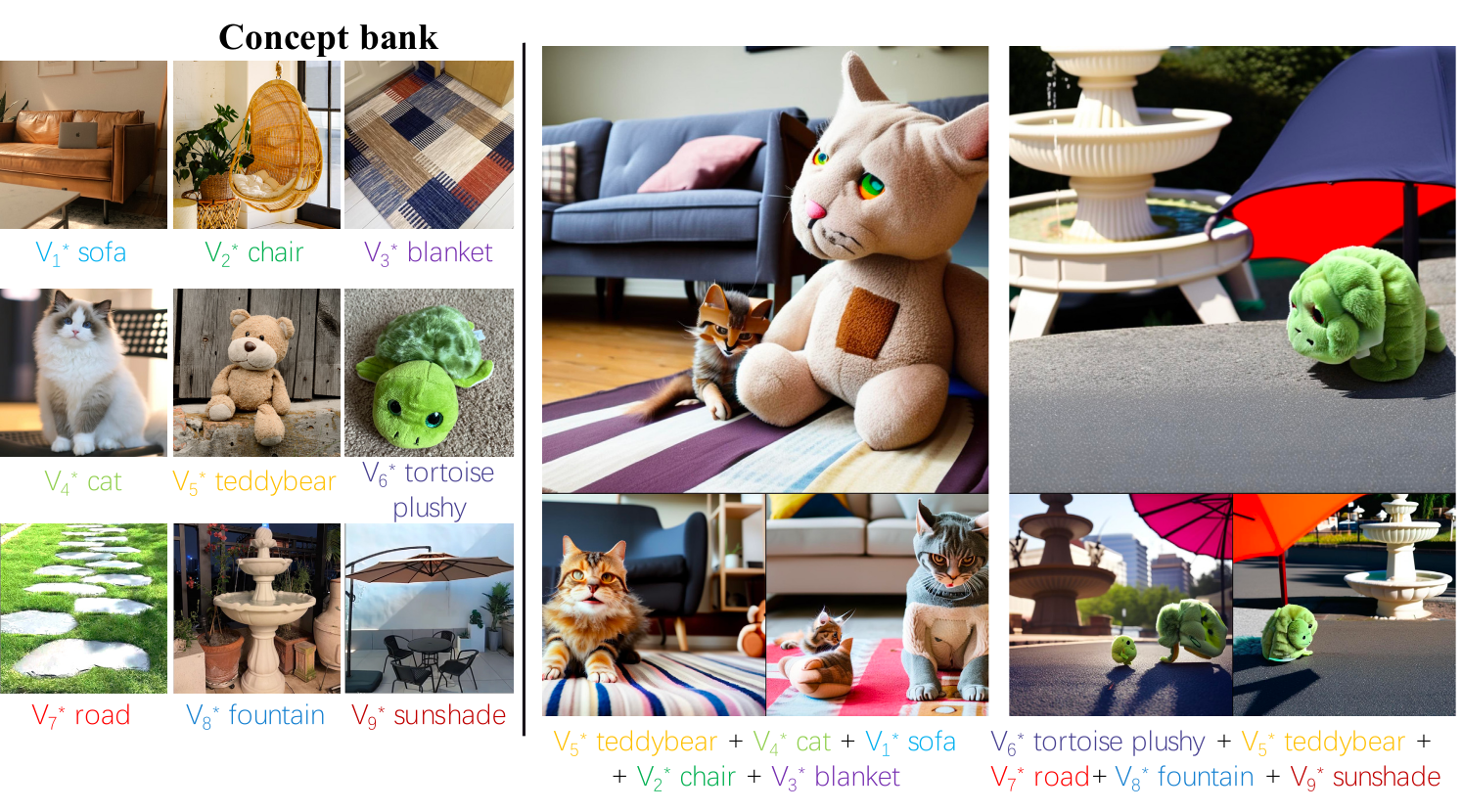}
 \caption{\textbf{Sample generations of Cones 2 with Stable Diffusion V2.1.}
 The generated images exhibit issues such as concept omission and low concept fidelity.
 }
 \label{fig:cones2_official}
\end{figure}

\paragraph{Cones 2.}
The official implementation\footnote{\url{https://github.com/ali-vilab/Cones-V2}} of Cones 2~\cite{liu2023cones2} uses Stable Diffusion V2.1 as its pre-trained diffusion model. 
For a fair comparison, we upgrade the backbone in the official implementation to SDXL.
We provide sample results generated by the original implementation in Fig.~\ref{fig:cones2_official} for reference. 
In the comparison of multi-concept generation without additional layout conditioning, explicit layout guidance is omitted.

\paragraph{OMG.}
We employ the official implementation\footnote{\url{https://github.com/kongzhecn/OMG}} of OMG~\cite{kong2024omg}.
The segmentation model $\texttt{SAM+Grounding-}$ $\texttt{DINO}$ is used to generate concept masks.
Following the authors' recommendation, we first train single-concept LoRA models using the code provided in the repository\footnote{\url{https://github.com/kohya-ss/sd-scripts}}.
To enhance the LoRA model's ability to capture key identifying features of concepts, we increase the rank dimensionality from 4 to 20.
During fine-tuning, the textual prompt for reference images is formatted as {\menlo "[V] <noun>"}.

\paragraph{MuDI.}
We utilize the official implementation\footnote{\url{https://github.com/agwmon/MuDI}} of MuDI~\cite{jang2024identity}. 
Following the authors' guidelines and examples, we make extra annotations for all reference images, including subject masks and detailed image captions.
The model is fine-tuned using the default parameters from the official implementation for a total of 2000 steps. 
Samples are generated using model checkpoints at steps 400, 600, 800, 1000, and 2000, with the best outputs selected for comparison.
For generation without layout conditions, we follow the official protocol,
employing latent initialization with random position.
For generation with explicit layout conditions, we manually specify the ordering and positioning of object initializations, incorporating masks as conditions into the initialization of latent variables.

\paragraph{\latexblend (Ours).}
Our code is implemented based on the diffusers library~\cite{vonplaten2022diffusers}.
Our method does not require additional annotations of reference images for fine-tuning.
The prompt template we used for fine-tuning is described in Section~\ref{temp_pool}.
We utilize real images as the regularization dataset, with a prior loss weight of 1.0.
The images are augmented using \texttt{RandomHorizontalFlip} with a flip probability of $0.5$.
The prompts for regularization images follow the format {\menlo "A <noun>"}.
The base learning rate for both model parameters and concept embeddings is set to $10^{-5}$.
After fine-tuning, concept representations are extracted from the prompt template {\menlo "Photo of \{\}."} for all concepts, as detailed in Section~\ref{pos_inv}.

\section{Societal Impact}\lblsec{societal}
Our method can seamlessly integrate multiple customized subjects into a single image with high quality while maintaining computational efficiency. This advancement democratizes access to high-quality text-to-image generation technologies, offering greater flexibility and personalization for customized content creation and enabling broader applications across creative industries.
However, the potential misuse of such technologies, including the generation of misleading or harmful content, raises ethical concerns. Recent research on safeguards, such as reliable detection methods for fake generated data, provides a promising approach to mitigating these potential negative impacts.

\clearpage

\end{document}